\documentclass[dvipsnames]{article}

\usepackage[final]{openbmb_2025}

\usepackage{amsmath}
\usepackage{amssymb}
\usepackage{mathtools}
\usepackage{amsthm}
\usepackage{multirow}
\usepackage{makecell}
\usepackage{caption}
\usepackage{subcaption}
\usepackage{wrapfig}
\usepackage{graphicx}
\usepackage{pifont}

\usepackage[utf8]{inputenc} % allow utf-8 input
\usepackage[T1]{fontenc}    % use 8-bit T1 fonts
\usepackage{url}            % simple URL typesetting
\usepackage{booktabs}       % professional-quality tables
\usepackage{amsfonts}       % blackboard math symbols
\usepackage{nicefrac}       % compact symbols for 1/2, etc.
\usepackage{microtype}      % microtypography

\usepackage{color, soul}
\usepackage[most,breakable]{tcolorbox}
\usepackage{xcolor}         % colors

\usepackage[labelfont=bf]{caption}
\usepackage{enumitem}
\usepackage{tablefootnote}
\usepackage{threeparttable}
\usepackage{tabularx}
\usepackage{booktabs}
\usepackage{colortbl} % 用于背景颜色，增加层次感
\definecolor{lightgray}{gray}{0.95}

\usepackage{fancyhdr}
\usepackage{wrapfig}
\renewcommand{\headrulewidth}{1pt}
\makeatletter
\def\headrule{{\if@fancyplain\let\headrulewidth\plainheadrulewidth\fi
\hrule\@height\headrulewidth\@width\textwidth \vskip-\headrulewidth}}
\makeatother

\definecolor{BMBDarkBlue}{HTML}{315EFE}
\definecolor{BMBLightBlue}{HTML}{00D3ED}

\usepackage[colorlinks, linkcolor=RoyalBlue, anchorcolor=RoyalBlue, citecolor=RoyalBlue, urlcolor=RoyalBlue]{hyperref}
\usepackage{amsthm}

\newcommand{\faHuggingFace}{%
  \raisebox{-0.13em}{%
    \includegraphics[height=1em]{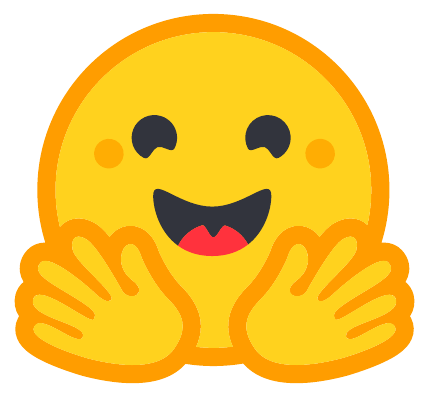}%
  }%
}

\newcommand{\faGithub}{%
  \raisebox{-0.13em}{%
    \includegraphics[height=1em]{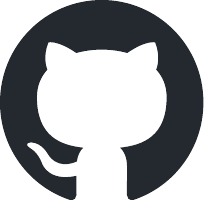}%
  }%
}

\newtcolorbox{mytheorem}{
  colback=gray!5, % 背景颜色
  colframe=gray!80, % 框线颜色
  boxrule=0.5pt, % 框线宽度
  arc=4pt, % 圆角半径
  left=4pt, % 左边距
  right=4pt, % 右边距
  top=4pt, % 上边距
  bottom=4pt, % 下边距
}

\title{Data Science and Technology Towards AGI \\ Part I: Tiered Data Management}

\author{%
Yudong Wang\textsuperscript{\rm 1}$^{*\dagger}$, 
Zixuan Fu\textsuperscript{\rm 1}$^{*}$, 
Hengyu Zhao\textsuperscript{\rm 2,3}$^{*}$, 
Chen Zhao\textsuperscript{\rm 2}$^{*}$, 
Chuyue Zhou\textsuperscript{\rm 2}$^{*}$, 
Xinle Lin\textsuperscript{\rm 2,4}$^{*}$, \\
\textbf{
Hongya Lyu\textsuperscript{\rm 2}, 
Shuaikang Xue\textsuperscript{\rm 2}, 
Yi Yi\textsuperscript{\rm 2}, 
Yingjiao Wang\textsuperscript{\rm 2}, 
Zhi Zheng\textsuperscript{\rm 2}, 
Yuzhou Zhang\textsuperscript{\rm 2}$^{\dagger}$, 
}\\
\textbf{
Jie Zhou\textsuperscript{\rm 2}$^{\dagger\ddagger}$, 
Chaojun Xiao\textsuperscript{\rm 1}$^{\ddagger}$, 
Xu Han\textsuperscript{\rm 1}$^{\ddagger}$, 
Zhiyuan Liu\textsuperscript{\rm 1}$^{\ddagger}$, 
Maosong Sun\textsuperscript{\rm 1}} \\
\textsuperscript{\rm 1}Tsinghua University ~~~~
\textsuperscript{\rm 2}ModelBest Inc. \\
\textsuperscript{\rm 3}Beijing Institute of Technology \\
\textsuperscript{\rm 4}South China Agricultural University \\
\texttt{yudongwang@tsinghua.edu.cn} \quad
\texttt{zhoujie@modelbest.cn} \quad
\texttt{\{xcj,han-xu,liuzy\}@tsinghua.edu.cn}
}

\newcommand\blfootnote[1]{%
\begingroup
\renewcommand\thefootnote{}\footnote{#1}%
\addtocounter{footnote}{-1}%
\endgroup
}

\begin{document}

\maketitle
\blfootnote{* Equal contribution.}
\blfootnote{$\ddagger$ Corresponding authors.}
\blfootnote{$\dagger$ Project leaders.}

\thispagestyle{fancy} % 恢复封面页的页眉和页脚

\vspace{-2em}

\begin{abstract}
The development of artificial intelligence can be viewed as an evolution of data-driven learning paradigms, with successive shifts in data organization and utilization continuously driving advances in model capability. Despite remarkable progress, current large language model (LLM) research is dominated by a paradigm that relies heavily on unidirectional scaling of data size, increasingly encountering bottlenecks in data availability, acquisition cost, and training efficiency.
In this work, we argue that the development of artificial general intelligence (AGI) is entering a new phase of data-model co-evolution, in which models actively guide data management while high-quality data, in turn, amplifies model capabilities. 
To implement this vision, we propose a tiered data management framework, designed to support the full LLM training lifecycle across heterogeneous learning objectives and cost constraints.
Specifically, we introduce an L0–L4 tiered data management framework, ranging from raw uncurated resources to organized and verifiable knowledge. Importantly, LLMs are fully used in data management processes, such as quality scoring and content editing, to refine data across tiers. Each tier is characterized by distinct data properties, management strategies, and training roles, enabling data to be strategically allocated across LLM training stages, including pre-training, mid-training, and alignment. The framework explicitly balances data quality, acquisition cost, and marginal training benefit, providing a systematic approach to scalable and sustainable data management.
We validate the effectiveness of the proposed framework through empirical studies on math and web data, in which tiered datasets are constructed from raw corpora and used across multiple training phases. Experimental results demonstrate that tier-aware data utilization significantly improves training efficiency and model performance. To facilitate further research, we release our tiered datasets and processing tools to the community.
\end{abstract}

\section{Introduction}
% 纵观人工智能的发展历程，本质上是一部“数据驱动策略与利用方式”的演进史。每一次范式跃迁既延伸和重构了前一阶段的数据驱动策略，又演进出新的数据利用方式，从而推动模型能力的跃升与涌现。
The development of artificial intelligence can be viewed as an evolution of data-driven strategies and data utilization paradigms \citep{zha2025data}. 
% 每一次范式跃迁既延伸和重构了前一阶段的数据驱动策略，又演进出新的数据利用和治理方式，从而推动模型能力的跃升与涌现。
Each paradigm shift extends and restructures prior approaches while introducing new methods for utilizing and managing data. These transformations have consistently driven improvements in model capability, enabling the emergence of higher-level intelligence \citep{wei2022emergent, gan2026beyond}.

% 根据核心数据类型的不同，我们可以将人工智能的发展分为四个阶段。
Based on the primary data types that drive each era, the developmental trajectory can be divided into four phases. \citep{buchanan1981dendral,shortliffe2012computer,rumelhart1986learning,cortes1995support,krizhevsky2012imagenet,he2016deep,vaswani2017attention}
% （1）符号学习：最初的符号智能时代确立了知识数据（Knowledge Data）驱动的范式，依赖专家将世界知识固化为静态的知识库（Knowledge Base），并通过显式规则实现智能化应用。这种直接的知识应用思路，在当前检索增强生成（Retrieval-Augmented Generation，RAG）技术的中得以延续。
(1) Symbolic Learning: the initial data era established a paradigm driven by knowledge data, such as human-annotated rules. It relied on experts to codify world knowledge into static knowledge bases \citep{augusto2021symbols} and implemented intelligence through explicit rules. % This philosophy of direct knowledge utilization maintains a clear conceptual lineage in modern Retrieval-Augmented Generation (RAG) technologies.
% （2）有监督学习：统计学习与深度学习时代开启了标注数据（Labeled Data）驱动的监督学习（Supervised Learning）范式，从人工特征工程演进到端到端的大规模监督训练，模型性能取决于数据规模、质量与表达能力的提升。在大模型时代，高质量标注数据的价值通过指令微调（Supervised Fine-Tuning，SFT）被重新定义，用于精确引导模型的输出行为。
(2) Supervised Learning: the era of statistical and deep learning established a paradigm driven by labeled data \citep{krizhevsky2012imagenet}. This stage witnessed the transition from manual feature engineering to end-to-end supervised training \citep{bengio2013representation, lecun2015deep}. The model performance became directly dependent on data scale, quality, and representational capacity \citep{sun2017revisiting}.  % Currently, the value of high-quality labeled data has been redefined through supervised Fine-Tuning (SFT), serving as a precision instrument to guide and align model behavior.
% （3）自监督学习：进入预训练时代，模型训练突破了对标注数据的完全依赖，进入以无监督数据（ Unsupervised Data）为驱动的自监督学习阶段，通过海量数据进行训练，使模型能够深度压缩和内化世界知识，从而带来了跨模态泛化能力与智能涌现。
(3) Self-supervised Learning: the pre-training era further reduced dependence on labeled data and enabled self-supervised learning driven by unsupervised data. Training on massive corpora allows models to compress and internalize world knowledge, leading to strong generalization and emergent capabilities across modalities \citep{wei2022emergent, achiam2023gpt, dubey2024llama}. 
% （4）反馈学习：随着进入反馈驱动时代，模型开始利用人类或环境反馈数据（Feedback Data）进行强化学习（Reinforcement Learning，RL），通过持续的交互和优化，实现对输出行为的主动探索和能力强化。这一阶段不仅提升了模型在复杂场景下的决策与适应能力，也为通用人工智能（Artificial General Intelligence，AGI）的演进奠定了基础。
(4) Feedback Learning: in the feedback-driven era, models leverage human and environmental feedback through reinforcement learning (RL) \citep{ziegler2019fine, kaufmann2024survey}. Continuous interaction enables active exploration of model behavior and capability improvement \citep{rafailov2023direct, liu2024deepseek}. This stage has strengthened decision-making and adaptability in complex settings and has laid an important foundation for progress toward artificial general intelligence (AGI) \citep{gan2026beyond}. 

\begin{figure}[!b]
    \centering
    \includegraphics[width=1.0\linewidth]{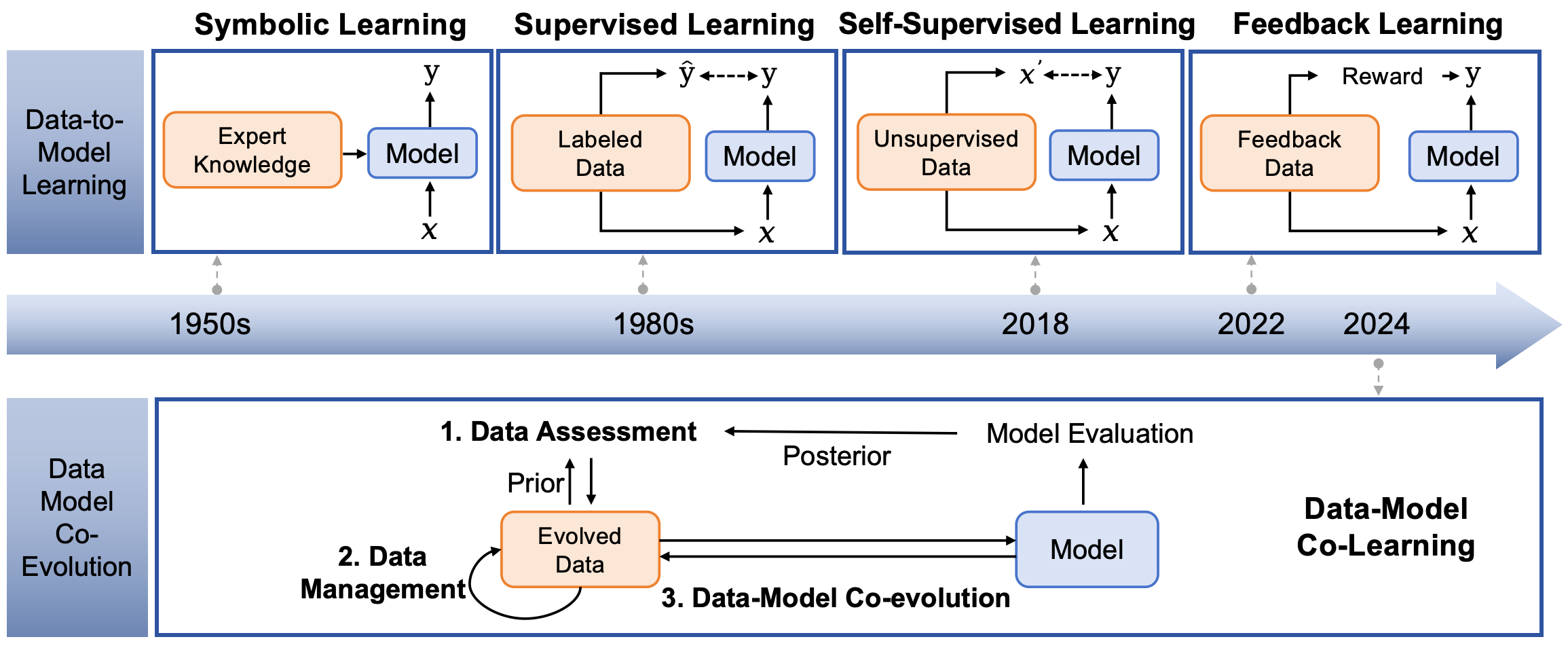}
    \caption{\textbf{Paradigm Shift in Data Organization and Utilization.} The evolution of LMs and ultimately toward AGI fundamentally represents a paradigm shift in the organization and utilization of data, progressing through Symbolic, Supervised, Self-supervised, and Feedback Learning phases. We argue that the field is transitioning toward a Data-Model Co-Learning phase, which necessitates three critical research pillars: Scientific Data Value Assessment, Hierarchical Data Management, and Dynamic Data-Model Co-evolution to transcend current sustainability bottlenecks.}
    \label{fig:1:evol_pipeline}
\end{figure}

As shown in Figure~\ref{fig:1:evol_pipeline}, current mainstream research primarily manifests as ``Data-Driven Learning'', which emphasizes unidirectional enhancement of model capabilities through expansion of data scale \citep{zhou2025survey}. As model capabilities advance, we argue that AI development should transition toward ``Data-Model Co-Evolution'', wherein models improve data management practices while high-quality data further refines model performance \citep{yuan2024self}, creating a positive feedback cycle. 

To accommodate this paradigm shift, this paper focuses on presenting a model-driven ``\textbf{Tiered Data Management}'' framework, aiming to provide systematic technical support for advancing toward artificial general intelligence.
The necessity of implementing tiered data management stems from three core considerations.
(1)~High-quality public data resources are becoming increasingly scarce. Future model development cannot rely solely on expanding data scale \citep{villalobos2022will}. Instead, data science and technology must shift from pursuing scale toward more careful data management and utilization.
(2)~LLM training involves multiple different phases -- from knowledge acquisition during pre-training to behavioral alignment \citep{ouyang2022training} during fine-tuning -- each with different requirements for data quality, quantity, and distribution \citep{mo2025mid,zha2025data}. This necessitates designing specialized training datasets suited to each phase's specific learning objectives.
(3)~Data management must balance the costs of data acquisition against the benefits to model performance. In the early stage of data management, lightweight and low-cost methods (such as heuristic filtering) should be adopted, while in deeper management stage, more fine-grained and higher-cost approaches (such as LLM-based labeling) should be used \citep{zhou2025survey}. Since high-quality data typically requires significant investment, strategically deploying valuable data at critical training moments -- such as mid-training phases or annealing stages -- can maximize data effectiveness while keeping overall costs manageable.

Despite significant academic progress in specific data processing tasks, such as filtering \citep{penedo2024fineweb, young2024yi, soldaini2024dolma}, selection \citep{chen2023alpagasus, penedo2023refinedweb, xie2023data, soldaini2024dolma, wettig2024qurating, engstrom2024dsdm, dubey2024llama}, and editing \citep{eldan2023tinystories, gunasekar2023textbooks, li2023textbooks, wang2023self, taori2023alpaca, peng2023instruction, xu2024wizardlm, wang2024codeclm, ding2023enhancing, cui2023ultrafeedback}, these approaches often fall short of addressing the systematic requirements of the LLM full-lifecycle training. To address this issue, we propose an L0-L4 tiered data management framework, evolving from raw resources to structured knowledge: 
(1)~\textbf{L0: Raw Data}. L0 data comprises PB-scale, uncurated resources characterized by high redundancy and noise, such as raw web dumps containing advertisements. It is maintained in its original state without deep processing. Consequently, it is primarily utilized for archiving and traceability rather than direct model training. 
(2)~\textbf{L1: Filtered Data}. L1 data features standardized text formatting and basic readability. It is usually produced via heuristic cleaning and deduplication to remove significant noise like web advertisements. As a result, it serves as the foundational resource pool for subsequent data selection and evaluation.
(3)~\textbf{L2: Selected Data}. L2 data retains samples with distinct themes and high information density, suitable for knowledge learning and domain adaptation (e.g., high-quality academic papers, technical code repositories, or filtered encyclopedia articles).
(4)~\textbf{L3: Refined Data}. L3 data features structured content with clear reasoning and explicit educational intent, ensuring maximum learnability. It is usually produced through rewriting, synthetic generation, or human refinement to achieve textbook-quality standards. Consequently, it serves as the core resource for advanced training phases like mid-training.
(5)~\textbf{L4: Organized Data}. L4 data consists of trustworthy and verifiable knowledge. It is created by converting unstructured text into organized formats, such as knowledge graphs or databases, and rigorously verifying the facts. Consequently, these data can provide the solid factual support necessary for retrieval-augmented generation.

To validate the effectiveness of the proposed tiered data management framework, we conduct a comprehensive empirical study across four representative domains, including English web, Chinese web, mathematics, and code data. 
By systematically constructing tiered datasets and applying them across the LLM training lifecycle, we demonstrate that the performance improves as data quality ascends from L1 to L3. 
Specifically, our results reveal that high-tier data (e.g., Math-L3) not only achieves domain-specific superiority but also acts as a fundamental driver for general reasoning, yielding significant cross-domain gains in language understanding and programming.
Furthermore, addressing the growing complexity of multi-stage training, we explore the application of tiered data management for multi-stage training to mitigate the interference of low-quality samples that often hampers late-stage convergence. 
Our analysis reveals that tiered training strategy, which introduces higher-quality data in the later phases, effectively prevents performance saturation and consistently outperforms mixed-training strategy.
These findings underscore the necessity of tiered data management as a core element of data science and technology for AGI, establishing the granular quality control essential for modern scaling laws.
As shown in Table~\ref{tab:ultradata_dataset_and_tools}, we have open-sourced an extensive collection of UltraData datasets and tools. Moving forward, we remain committed to the continuous release of these resources and invite the community to integrate our tiered management logic into the evolving landscape of data science and technology towards AGI.

\begin{table}[!h]
	% \footnotesize  %
	\setlength{\tabcolsep}{9pt}
	\centering
    \small
    \caption{Open-source datasets and tools released in this paper.}
	\begin{tabular}{lllrc}
		\toprule
		\textbf{Type} & \textbf{Name} & \textbf{Description} & \textbf{Scale} & \textbf{Link} \\
		\midrule
        % \multicolumn{4}{c}{Dataset} \\
        % \midrule
        \multirow{7}{*}{Dataset} &
		UltraData-Math-L1 & Filtered math data using heuristic rules & 170B & \href{https://github.com/UltraData-OpenBMB/UltraData-Math/}{\faGithub}\;\href{https://huggingface.co/datasets/openbmb/UltraData-Math}{\faHuggingFace} \\
        & UltraData-Math-L2 & Model-selected math data & 33B & \href{https://github.com/UltraData-OpenBMB/UltraData-Math/}{\faGithub}\;\href{https://huggingface.co/datasets/openbmb/UltraData-Math}{\faHuggingFace} \\
		& UltraData-Math-L3 & Synthetic and refined math data & 88B & \href{https://github.com/UltraData-OpenBMB/UltraData-Math/}{\faGithub}\;\href{https://huggingface.co/datasets/openbmb/UltraData-Math}{\faHuggingFace} \\ \cmidrule{2-5}
		& Ultra-Fineweb-en (L2) & Model-selected English web corpus & 1,800B & \href{https://github.com/openbmb/minicpm}{\faGithub}\;\href{https://huggingface.co/datasets/openbmb/Ultra-FineWeb}{\faHuggingFace} \\	
        & Ultra-Fineweb-en-L3 & Synthetic and refined English web data & 200B & \href{https://github.com/openbmb/minicpm}{\faGithub}\;\href{https://huggingface.co/datasets/openbmb/Ultra-FineWeb-L3}{\faHuggingFace} \\	\cmidrule{2-5}
		& Ultra-Fineweb-zh (L2) & Model-selected Chinese web corpus & 120B & \href{https://github.com/openbmb/minicpm}{\faGithub}\;\href{https://huggingface.co/datasets/openbmb/Ultra-FineWeb}{\faHuggingFace} \\
		& Ultra-Fineweb-zh-L3 & Synthetic and refined Chinese web data & 200B & \href{https://github.com/openbmb/minicpm}{\faGithub}\;\href{https://huggingface.co/datasets/openbmb/Ultra-FineWeb-L3}{\faHuggingFace} \\	
        \midrule
        % \multicolumn{4}{c}{Tool} \\
        % \midrule
        \multirow{4}{*}{Tool}
		& UltraData-Math-Parser & Enhanced HTML parser for math content & / & \href{https://github.com/UltraData-OpenBMB/UltraData-Math/tree/main/UltraData-Math-L0-Parser}{\faGithub}\;\href{https://huggingface.co/spaces/openbmb/UltraData-Math-L0-Parser}{\faHuggingFace} \\	
		& UltraData-Math-Generator & Synthetic math problem generator & / & \href{https://github.com/UltraData-OpenBMB/UltraData-Math/tree/main/UltraData-Math-L3-Generator}{\faGithub}\;\href{https://huggingface.co/spaces/openbmb/UltraData-Math-L3-Generator}{\faHuggingFace} \\	\cmidrule{2-5}
		& Ultra-FineWeb-en-Classifier & Classifier for selecting English web data & / & \href{https://github.com/openbmb/minicpm}{\faGithub}\;\href{https://huggingface.co/openbmb/Ultra-FineWeb-classifier}{\faHuggingFace} \\	
		& Ultra-FineWeb-zh-Classifier & Classifier for selecting Chinese web data & / & \href{https://github.com/openbmb/minicpm}{\faGithub}\;\href{https://huggingface.co/openbmb/Ultra-FineWeb-classifier}{\faHuggingFace} \\	
		\bottomrule
	\end{tabular}
	\label{tab:ultradata_dataset_and_tools}
\end{table}

\section{Tiered Data Management Framework}

% 大模型训练中，数据质量是决定模型性能上限的关键要素。
In the training of LLMs, data quality serves as textbooks for model learning. 
% 虽然数据治理已吸引了许多学者的努力，但现有工作尚未明确提出模型的分级治理体系。
% Despite extensive research in data management, a systematic hierarchical framework remains absent.
% 当前大模型构建中，不同质量的数据被粗放地混合在一起进行训练，使得高价值数据难以被精准学习与利用，造成数据利用效率低。
In recent LLM construction, data of varying quality are often indiscriminately mixed during training. This coarse-grained approach hinders the fully utilization of high-value data samples, leading to suboptimal model performance. 
% 已有研究表明，高质量数据在模型退火阶段、Mid-training更多地使用，能够显著提升模型性能\citep{wang2025octothinker}。
Prior studies have demonstrated that concentrating high-quality data during specific phases, such as annealing or mid-training, can significantly enhance model performance~\citep{minicpm,wang2025octothinker}.
% 本章首先从数据类型与处理方法两个维度对既有研究进行总结，然后提出一套基于数据质量梯度的精细化分级治理体系，实现数据在采集、清洗、筛选与使用各环节的有序衔接与动态优化。
In this section, we first review existing data management research from the perspectives of training stages and processing methodologies. Building on this summary, we propose a fine-grained management framework with data quality-based tiering. This framework facilitates seamless transitions and dynamic optimization across the entire data lifecycle, encompassing collection, cleaning, filtering, and utilization.

% 2.2.1 现有数据分级治理体系
% 当前的治理体系主要依据模型训练阶段（如预训练、微调）或具体处理方法（如过滤、筛选）进行划分。
\subsection{Existing Data Management Frameworks}
Existing data management frameworks are typically organized by model training stages or specific data processing methodologies. In this section, we first introduce the stage-oriented management framework and the method-oriented management framework. Then we propose our tiered data management framework.

% （1）基于模型训练阶段的治理体系
% 这种治理体系深度耦合模型的全生命周期训练流程，依据“知识获取”、“领域强化”与“能力对齐”等不同目标，将数据划分为服务于预训练、领域适应、后训练等不同阶段的专用类型。针对不同训练阶段对数据规模、多样性及信噪比的需求差异显著，该体系通常构建并行的治理标准与差异化的处理管线。
\subsubsection{Stage-Oriented Management Framework}
This framework is tightly coupled with the entire model training lifecycle. Driven by distinct objectives, such as knowledge acquisition, domain enhancement, and capability alignment, it categorizes corpora and methods into specialized categories for pre-training, mid-training, and post-training phases. Given the significant variations in requirements for scale, diversity, and signal-to-noise ratio across these stages, the framework typically establishs parallel management standards and differentiated processing pipelines.

% - 预训练数据治理：预训练数据治理旨在在海量数据规模与高有效信息密度之间寻求最优平衡，同时确保领域覆盖的多样性，其核心是赋予模型广泛的通识理解能力与稳健的语言表征基础。随着大模型技术的不断发展，数据治理范式经历了从早期的启发式规则过滤，到统计去重与模型驱动筛选，再到主动合成与生成的演变。
\textbf{Pre-training data management.} 
Pre-training management strives for an optimal equilibrium between data scale and information density, while strictly preserving domain diversity. Its core objective is to endow models with broad general knowledge and establish a solid foundation for linguistic representation. As data engineering techniques evolve, management paradigms are transitioning from early heuristic rule-based filtering to statistical deduplication and model-driven selection, and ultimately advancing toward active synthesis and generation. 
% 早期工作如 C4\citep{raffel2020exploring} 确立了 Web 数据清洗的基线，主要依赖启发式规则和语言识别剔除低质文本。随后，RefinedWeb\citep{penedo2023refinedweb} 证明通过大规模模糊去重和严格 URL 过滤，纯 Web 数据即可超越 The Pile\citep{gao2020pile} 等混合语料。FineWeb-Edu\citep{penedo2024fineweb}、DCLM\citep{li2024datacomp} 和 Ultra-FineWeb\citep{wang2025ultra} 利用模型分类器对网页进行“教育价值”打分，证明了基于强模型的筛选显著优于传统启发式规则。在高质量语料稀缺的背景下，Phi\citep{gunasekar2023textbooks, abdin2024phi} 开创了“教科书级”合成与训练数据的范式，Nemotron-CC\citep{su2025nemotron} 进一步扩充了预训练合成的范式，利用强模型生成无噪声语料。
Foundational efforts such as C4 \citep{raffel2020exploring} established the baseline for web data cleaning, relying predominantly on heuristic rules and language identification to eliminate low-quality text. Building on this, RefinedWeb \citep{penedo2023refinedweb} demonstrated that large-scale fuzzy deduplication and strict URL filtering enable pure web corpora to surpass mixed datasets like The Pile \citep{gao2020pile}. Moreover, FineWeb-Edu \citep{penedo2024fineweb}, DCLM \citep{li2024datacomp}, and Ultra-FineWeb \citep{wang2025ultra} employed model-based classifiers to score webpages for educational value, providing compelling evidence that filtering driven by powerful models significantly outperforms traditional heuristics. In response to the impending scarcity of high-quality natural corpora, the Phi series \citep{gunasekar2023textbooks, abdin2024phi} introduced a textbook-level synthetic paradigm, while Nemotron-CC \citep{su2025nemotron} expanded this frontier by leveraging strong models to generate nearly ``noise-free'' pre-training corpora.

% - 领域数据治理：领域数据治理更侧重于专业知识和逻辑推理等领域数据，旨在提模型在领域任务上的能力。
\textbf{Mid-training data management.} Mid-training management prioritizes specialized knowledge and structured reasoning to bolster model performance on vertical tasks. 
% 数学数据的治理核心在于从非结构化网页中还原严密的逻辑结构。OpenWebMath\citep{paster2023openwebmath} 和 MegaMath\citep{zhou2025megamath} 通过优化 HTML 解析策略，显著提升了网页中 Latex 公式的解析能力，有效解决了传统提取中公式丢失或错位的问题。为了筛选通用网页中的高质量数学内容，MathScore\citep{paster2023openwebmath}、Deepseek-Math\citep{shao2024deepseekmath} 等训练了领域专属分类器进行语义级筛选，确保数据的专业纯度。在此基础上，Nemotron-CC-Math\citep{mahabadi2025nemotron} 进一步提出了“解析+编辑”的范式，先利用 Lynx 渲染保留二维结构，再利用大模型作为“编辑器”对解析后的文本进行清洗与重写，修复断裂的推理步骤，构建出高保真的无噪声语料。
For mathematical data, the primary challenge involves reconstructing rigorous logical structures from unstructured web content. OpenWebMath (OWM) \citep{paster2023openwebmath} and MegaMath \citep{zhou2025megamath} have substantially enhanced LaTeX formula extraction via optimized HTML parsing strategies, effectively mitigating issues of formula omission and misalignment that are prevalent in traditional extraction methods. To identify and retain high-quality mathematical content, MathScore \citep{paster2023openwebmath} and DeepSeek-Math \citep{shao2024deepseekmath} develop domain-specific classifiers for semantic-level filtering. Building on these foundations, Nemotron-CC-Math \citep{mahabadi2025nemotron} pioneered a ``parsing-then-editing'' paradigm. By preserving the visual layout of mathematical expressions through Lynx rendering and employing large language models (LLMs) as neural editors, this approach repairs fragmented reasoning steps, yielding high-fidelity, noise-free corpora. 
% 在代码数据上，The Stack v2\citep{lozhkov2024starcoder} 采用了一套严格的启发式过滤与去重管线，通过限制代码行长、字母数字比例等规则剔除低质代码，并利用 MinHash 算法去除大规模重复内容，为模型提供了纯净的训练底座。在此基础上，DeepSeek-Coder-V2\citep{zhu2024deepseek} 进一步引入了模型驱动的筛选机制（Model-Based Filtering），利用从高质量种子数据训练的分类器，并从 Common Crawl 中精准召回被遗漏的代码与技术文档，显著拓展了数据的来源。针对代码注释稀缺的问题，Code Needs Comments\citep{song2024code} 和 AlchemistCoder\citep{song2024alchemistcoder} 采用合成增强策略，利用 LLM 为裸代码生成详细文档或合成编程问题，显著提升了代码数据的质量。
In the realm of code data, The Stack v2 \citep{lozhkov2024starcoder} establishes a robust baseline by adopting strict heuristic filtering and deduplication pipelines. DeepSeek-Coder-V2 \citep{zhu2024deepseek} expands upon this by incorporating model-based filtering to retrieve high-quality code and technical documentation previously overlooked in Common Crawl. Furthermore, synthetic augmentation strategies such as Code Needs Comments \citep{song2024code} and AlchemistCoder \citep{song2024alchemistcoder} significantly refine code quality by generating detailed documentation and synthesizing programming tasks.
% 此外，在法律领域（如 SaulLM\citep{colombo2024saullm}）与医疗领域（如 PMC-LLaMA\citep{wu2024pmc}）等垂直场景中，研究者们设计了基于引用的去重策略与基于实体抽取的过滤机制，以确保知识的溯源性与准确性。
Beyond general coding tasks, vertical domains such as law and medicine require more stringent management. Works such as SaulLM~\citep{colombo2024saullm} and PMC-LLaMA \citep{wu2024pmc} have engineered citation-based deduplication strategies and entity-extraction-based filtering mechanisms to guarantee knowledge traceability and factual accuracy.

% - 后训练数据治理：后训练数据治理关注基础模型预训练完成后在指令微调和强化学习阶段对数据进行管理和筛选，以确保模型更好地响应人类指令并遵循预期行为准则。这一阶段的数据治理同样强调数据质量与多样性的平衡，广泛借助强模型合成与定制化筛选策略实现精细化管理。
\textbf{Post-training data management.} 
Post-training data management focuses on curating and filtering data during the instruction fine-tuning and reinforcement learning stages after base model pre-training, aiming to improve the model's responsiveness to human instructions and adherence to intended behavioral norms. This stage also emphasizes balancing data quality and diversity, extensively leveraging strong-model synthesis and customized filtering strategies to achieve fine-grained control.
% 在指令微调阶段，典型的工作包括 Self-Instruct\citep{wang2023self}，建立了利用少量人工种子任务引导模型自我生成数据的基本范式；Evol-Instruct\citep{xu2023wizardlm} 通过深度与广度进化机制系统性提升数据的逻辑难度与覆盖范围；OSS-Instruct\citep{wei2023magicoder} 引入外部开源代码作为先验约束，有效缓解合成数据分布偏差；而 Magpie\citep{xu2024magpie} 则利用对齐模型的自回归特性，实现了无种子的“内省式”指令生成，挖掘模型潜在能力空间。
During the instruction/supervised fine-tuning (SFT) phase, representative methodologies have focused on scaling and refining synthetic instructions. Self-Instruct \citep{wang2023self} established the foundational paradigm of bootstrapping instruction data by guiding models with a minimal set of human-authored seed tasks. UltraChat \citep{ding2023enhancing} extended this paradigm to multi-turn conversational settings, employing separate LLMs to iteratively simulate user and assistant roles and systematically covering diverse thematic sectors including world knowledge, creative writing, and document-grounded assistance. Evol-Instruct \citep{xu2023wizardlm} advanced this by systematically escalating logical difficulty and coverage through evolutionary mechanisms along both depth and breadth dimensions. To address distributional bias, OSS-Instruct \citep{wei2023magicoder} introduced external open-source code as a high-quality logical prior. More recently, Magpie \citep{xu2024magpie} exploited the autoregressive properties of aligned models to facilitate seed-free, introspective instruction generation, effectively uncovering latent capability spaces without external reliance. 
% 在数据筛选方面，受 LIMA\citep{zhou2023lima} “少即是多”原则启发，研究者转向模型驱动策略以提升指令信息密度：MoDS\citep{du2023mods}与DEITA\citep{liu2023makes}分别从质量、覆盖度及复杂度维度量化样本价值。
In terms of SFT data selection, inspired by the ``less is more'' principle advocated by LIMA \citep{zhou2023lima}, research has pivoted toward model-driven strategies to maximize the information density of instruction data. Approaches such as MoDS \citep{du2023mods} and DEITA \citep{liu2023makes} quantify data value by rigorously evaluating samples across quality, coverage, and complexity dimensions. 
% 在强化学习阶段，治理重心转向偏好信号的精准构建，UltraFeedback\citep{cui2023ultrafeedback} 结合多模型采样与多维度 AI 细粒度评分（如指令遵循、真实性等）的复杂对齐体系。近期，强化学习数据正从强调“风格偏好”转向侧重“逻辑正确性”，在数学、代码等领域，逐渐转向构建“问题 + 可验证答案”的闭环，通过奖励函数（Reward Function）对结果进行判别。这种方式减少了对过程标注的依赖，使模型在反馈循环中通过自主探索实现能力的持续演进与跃迁。
In the reinforcement learning (RL) stage, data management mainly focuses on the precise construction of preference signals. UltraFeedback \citep{cui2023ultrafeedback} pioneered this approach by coupling multi-model sampling with fine-grained, multi-dimensional AI annotations. By evaluating criteria such as instruction following and honesty, this framework synthesizes high-quality alignment signals to refine model performance. More recently, RL data paradigms have evolved from emphasizing stylistic preferences to prioritizing logical correctness. In reasoning-intensive domains such as mathematics and coding, this trend manifests in the construction of closed-loop datasets featuring problems with verifiable answers, where deterministic reward functions are employed to adjudicate outcomes \citep{shao2024deepseekmath, guo2025deepseek}. This methodology reduces reliance on expensive process-level annotations \citep{lightman2023let}, enabling models to achieve sustained capability evolution and qualitative leaps through autonomous exploration within feedback loops.

% （2）基于数据处理方法的分级治理体系
\subsubsection{Method-Oriented Management Framework}

% 这种治理体系侧重于数据处理方法。它不以训练阶段为界限，而是根据处理逻辑的复杂程度与介入智能的深度，构建了一套从基础清洗到高级合成的流水线范式。
% This framework emphasizes the processing methods. 
This framework centers on processing methodologies. Rather than being delineated by training stages, it organizes the data pipeline according to the complexity of processing logic and the depth of intelligence involved, forming a hierarchical workflow that ranges from basic cleaning to advanced data synthesis.

\textbf{Data Parsing.} As the entry point of the management pipeline, data parsing transforms heterogeneous raw files into machine-interpretable, logically coherent structured assets.
For HTML parsing, traditional tools like Trafilatura \citep{barbaresi-2021-trafilatura} employ heuristic rule-based matching algorithms to separate meaningful content from boilerplate elements. Recent advances reformulate content extraction as a semantic understanding problem, leveraging language models to perform sequence labeling on DOM structures (e.g., MinerU-HTML \citep{liu2025drippertokenefficientmainhtml}) or transform HTML into structured Markdown and JSON formats (e.g., ReaderLM-v2 \citep{wang2025readerlm}). For mathematical content, specialized approaches such as OpenWebMath \citep{paster2023openwebmath}, MegaMath \citep{zhou2025megamath}, and Nemotron-CC-Math \citep{mahabadi2025nemotron} achieve precise LaTeX formula extraction through optimized HTML processing strategies (detailed in the mid-training management section).
For document parsing, pipelines rely on customized frameworks such as MinerU \citep{wang2024mineru} and PaddleOCR \citep{cui2025paddleocr}, end-to-end models such as Nougat \citep{blecher2023nougat}, or OCR solutions powered by Vision-Language Models (e.g., GOT-OCR \citep{wei2024general}, olmOCR \citep{poznanski2025olmocr}, DeepSeek-OCR~\citep{wei2026deepseek}, Qwen3-VL~\citep{bai2025qwen3vltechnicalreport}, and DeepSeek-OCR2~\citep{zhong2026ocrverse}) to transform complex-layout documents into structured formats.
For audio parsing, models such as Whisper \citep{radford2023robust} enable high-fidelity transcription, converting audio streams into timestamped, speaker-diarized transcripts.    

% - 数据过滤（Data Filtering）： 数据过滤是治理体系的基石，旨在通过低成本的工程化手段剔除噪声从而保证数据质量的下限。
\textbf{Data Filtering.} Data filtering serves as the bedrock of the management pipeline, aiming to eliminate noise through low-cost, engineering-centric mechanisms and thereby ensure a minimum quality threshold.  
% 早期以 C4\citep{raffel2020exploring} 为代表的范式主要依赖启发式规则，通过正则表达式、关键词列表及语言识别模型（Language ID）剔除低质、重复或非目标语言内容。随着对数据理解的深入，RefinedWeb\citep{penedo2023refinedweb} 进一步强化了大规模模糊去重（Fuzzy Deduplication）与严格的 URL 过滤机制，证明了高质量过滤能显著提升预训练的起步效率。在多模态领域，数据过滤还涉及图像文本匹配度评分（如 CLIP Score）等跨模态清洗手段，以确保基础关联的准确性。
Early paradigms, exemplified by C4 \citep{raffel2020exploring}, relied predominantly on heuristic rules. These methods utilize regular expressions, blacklists, and language identification models to excise low-quality, duplicated, or non-target language content. Building upon these foundations, RefinedWeb \citep{penedo2023refinedweb} advanced this by implementing large-scale fuzzy deduplication via MinHash-LSH \citep{broder1997resemblance} and rigorous URL filtering, demonstrating that superior filtering significantly boosts pre-training efficiency. Beyond surface-level matching, SemDeDup \citep{abbas2023semdedup} leverages embeddings to identify semantic duplicates, enabling more aggressive data reduction with minimal performance loss. In multimodal contexts, filtering extends to cross-modal alignment techniques, such as image-text matching scoring (e.g., CLIP Score), ensuring precise semantic correspondence between modalities.

\textbf{Data Selection.} Data selection utilizes model-based classifiers to perform multi-dimensional filtering based on data quality and thematic relevance, often enriching samples with semantic annotations in the process.
DCLM \citep{li2024datacomp}, FineWeb-Edu \citep{penedo2024fineweb}, and Ultra-FineWeb \citep{wang2025ultra} employ classifiers to identify and prioritize corpora with high educational value. Similarly, DeepSeek-Math \citep{shao2024deepseekmath} and FineMath \citep{allal2025smollm2} construct specialized classifiers to pinpoint high-quality, reasoning-intensive samples in the mathematical domain.
In the post-training stage, methods such as DEITA \citep{liu2023makes} and MoDS \citep{du2023mods} assess data utility across dimensions of quality, coverage, and necessity, employing optimization-driven algorithms or influence functions to distill massive instruction pools into compact yet highly representative subsets.
Beyond binary quality filtering, DecorateLM \citep{zhao2024decoratelm} distills expert knowledge from large models into lightweight annotators, enabling efficient processing of hundreds of billions of tokens. Its three-level tagging system automatically appends hierarchical knowledge labels and performs standardized editing on raw text, significantly enhancing the structural quality of pre-training corpora. QuRating \citep{wettig2024qurating} further advances quality assessment by eliciting pairwise comparisons from LLMs across dimensions such as writing style, required expertise, and educational value, training scalar rating models that enable fine-grained, continuous-valued data selection at scale. At a finer granularity, Rho-1 \citep{lin2024rho} introduces Selective Language Modeling that operates at the token level, dynamically focusing training on high-value tokens while skipping uninformative ones.

% - 数据编辑（Data Editing）： 数据编辑是对现有存量数据的精化与重构，旨在修复数据瑕疵并增强逻辑连贯性。
% \textbf{Data Editing.} 
% Data editing focuses on the refinement and restructuring of existing datasets to rectify defects and enhance logical coherence.  
% % 针对原始网页解析后存在的公式断裂或格式错位，Nemotron-CC-Math\citep{mahabadi2025nemotron} 直接对解析后的数学数据进行了编辑。此外， Qwen3\citep{yang2025qwen3} 针对文档数据，在 VLM 解析后也使用 LLM 对数据进行了编辑。此外，ProX\citep{zhou2024programming} 与 RefineX\citep{bi2025refinex} 提出了编程方式编辑数据的范式，将数据精炼视为编程任务，让模型自动生成细粒度的编辑操作（如字符串规范化、噪声删除）来逐条优化语料。
% To address specific issues such as formula fragmentation and formatting inconsistencies inherent in web parsing, Nemotron-CC-Math \citep{mahabadi2025nemotron} employs direct editing of parsed mathematical data. Similarly, Qwen3 \citep{yang2025qwen3} integrates LLM-based editing post-VLM parsing to optimize document processing. Furthermore, ProX \citep{zhou2024programming} and RefineX \citep{bi2025refinex} conceptualize data refinement as a programming task. These frameworks empower models to autonomously generate fine-grained editing operations (\textit{e.g.}, string normalization and noise removal) to iteratively polish corpus quality.

\textbf{Data Editing.}
Unlike synthesis, which generates new content, data editing refines and restructures existing datasets to rectify defects and enhance logical coherence.
ProX \citep{zhou2024programming} and RefineX \citep{bi2025refinex} formulate data refinement as a programming task, enabling models to autonomously generate fine-grained editing operations (\textit{e.g.}, string normalization and noise removal) for iterative corpus polishing. Additionally, Nemotron-CC-Math \citep{mahabadi2025nemotron} and Qwen3 \citep{yang2025qwen3} employ LLM-based editing to repair formula fragmentation and formatting inconsistencies in parsed documents.

% - 数据合成（Data Synthesis）： 数通过模型生成实现高质量数据的规模化供给，主要包括预训练合成和后训练合成两种类型。
% \textbf{Data Synthesis.}
% Data synthesis facilitates the scalable production of high-quality data via generative models, spanning both pre-training and post-training phases. 
% % Phi 系列的“教科书级”语料合成开启了预训练合成的范式。Self-Instruct\citep{wang2023self} 与 Evol-Instruct\citep{xu2023wizardlm} 的指令自动演进，合成技术已从简单的模仿转向复杂的逻辑演变。OSS-Instruct\citep{wei2023magicoder} 展示了如何利用开源代码作为“逻辑先验”引导合成，而 Magpie\citep{xu2024magpie} 则挖掘了模型的自回归潜力，实现了内省式的零成本指令生成。
% The textbook-level corpus generation pioneered by the Phi series \citep{gunasekar2023textbooks} inaugurated a new paradigm for pre-training synthesis. Nemotron-CC \citep{su2025nemotron} further expanded this frontier by leveraging strong teacher models to synthesize large-scale, low-noise pre-training corpora. In the realm of instruction tuning, Self-Instruct \citep{wang2023self} and Evol-Instruct \citep{xu2023wizardlm} advanced synthesis through automated evolution, shifting the focus from simple imitation to complex logical construction. Furthermore, OSS-Instruct \citep{wei2023magicoder} demonstrated the utility of open-source code as a logical prior for guided synthesis, while Magpie \citep{xu2024magpie} exploited autoregressive capabilities to achieve introspective, near-zero-cost instruction generation.

\textbf{Data Synthesis.}
Data synthesis enables scalable production of high-quality data via generative models, spanning both pre-training and post-training phases.
For pre-training, the Phi series \citep{gunasekar2023textbooks} pioneered the textbook-level corpus generation paradigm, while Nemotron-CC \citep{su2025nemotron} further scaled this approach by leveraging strong teacher models to synthesize large-scale, low-noise corpora.
For post-training, instruction synthesis has evolved from simple imitation to complex logical construction through methods such as Self-Instruct \citep{wang2023self}, Evol-Instruct \citep{xu2023wizardlm}, OSS-Instruct \citep{wei2023magicoder}, and Magpie \citep{xu2024magpie}, as detailed in the post-training data management section.

% 综上所述，尽管方法多样，但现有实践在分级与系统化管理上仍存在显著短板：缺乏统一的分级规范，导致高价值数据难以凸显；治理策略单一，难以根据数据价值、用途或模型阶段实施差异化处理；数据处理流程割裂严重，采集、清洗、筛选与验证各环节往往独立执行，缺乏统一的质量反馈与闭环机制。同时，现有方法难以实现数据血缘追踪，导致数据来源、处理路径及修改记录难以追溯，进一步阻碍了流程的自动化与智能化升级。
% In summary, despite the diversity of these methodologies, current data management suffers from significant structural and systemic deficiencies. 
% Primarily, the absence of unified hierarchical standards obscures the effective identification and prioritization of high-value data. Second, monolithic management strategies fail to accommodate the nuanced requirements of different data types or model development/training stages. Data processing pipelines are highly fragmented, with data collection, cleaning, filtering, and validation often conducted in isolation, lacking a unified quality assessment standard and a closed-loop feedback mechanism. Moreover, existing approaches struggle to support data lineage tracking, making it difficult to trace data provenance, processing paths, and modification histories, which further hinders the automation and intelligent upgrading of data management workflows.

In summary, despite methodological diversity, existing practices remain hampered by the lack of systematic and hierarchical management. The absence of unified tiered standards makes it difficult to identify and prioritize high-value data effectively. Management strategies tend to be monolithic, failing to implement differentiated processing based on data value, intended use, or training stage. Moreover, data processing pipelines are severely fragmented—collection, cleaning, selection, and validation are often conducted independently without unified quality metrics or closed-loop feedback mechanisms. Additionally, current approaches provide inadequate support for data lineage tracking, making it difficult to trace data provenance, processing paths, and modification histories, which further impedes the automation and intelligent evolution of data management workflows.

\subsection{Tiered Data Management Framework}
% 2.2.2 精细化数据分级治理体系 
% \subsubsection{L0-L4 Hierarchical Framework}
% 为了突破现有治理体系的局限，必须建立一套以“价值密度”和“可信度”为核心的精细化分级标准。因此，我们提出 L0-L4 数据分级体系，旨在通过标准化的层级定义，实现数据资产的有序管理与高效流转。该体系借鉴自动驾驶分级理念，根据数据的信息密度、加工程度等信息，将数据划分为五个等级。每一级都代表了更高的数据纯度，同时也对应着更高的获取成本。
To overcome the limitations of existing management frameworks, it is essential to establish a fine-grained hierarchical standard centered on data quality and trustworthiness. Specifically, we define five distinct levels (L0--L4). Each level represents a progressive increase in data purity, albeit with a corresponding rise in acquisition and computational costs. In what follows, we elaborate on the specific definitions and representative datasets for each level, while providing typical case studies based on web and math data to illustrate their practical implementation. Furthermore, to provide a concrete and actionable reference, we systematically organize representative open-source tools and publicly available datasets according to our tiered criteria, as summarized in Table~\ref{tab:data_tier_system}.

\begin{figure}[!t]
    \centering
    \includegraphics[width=0.7\linewidth]{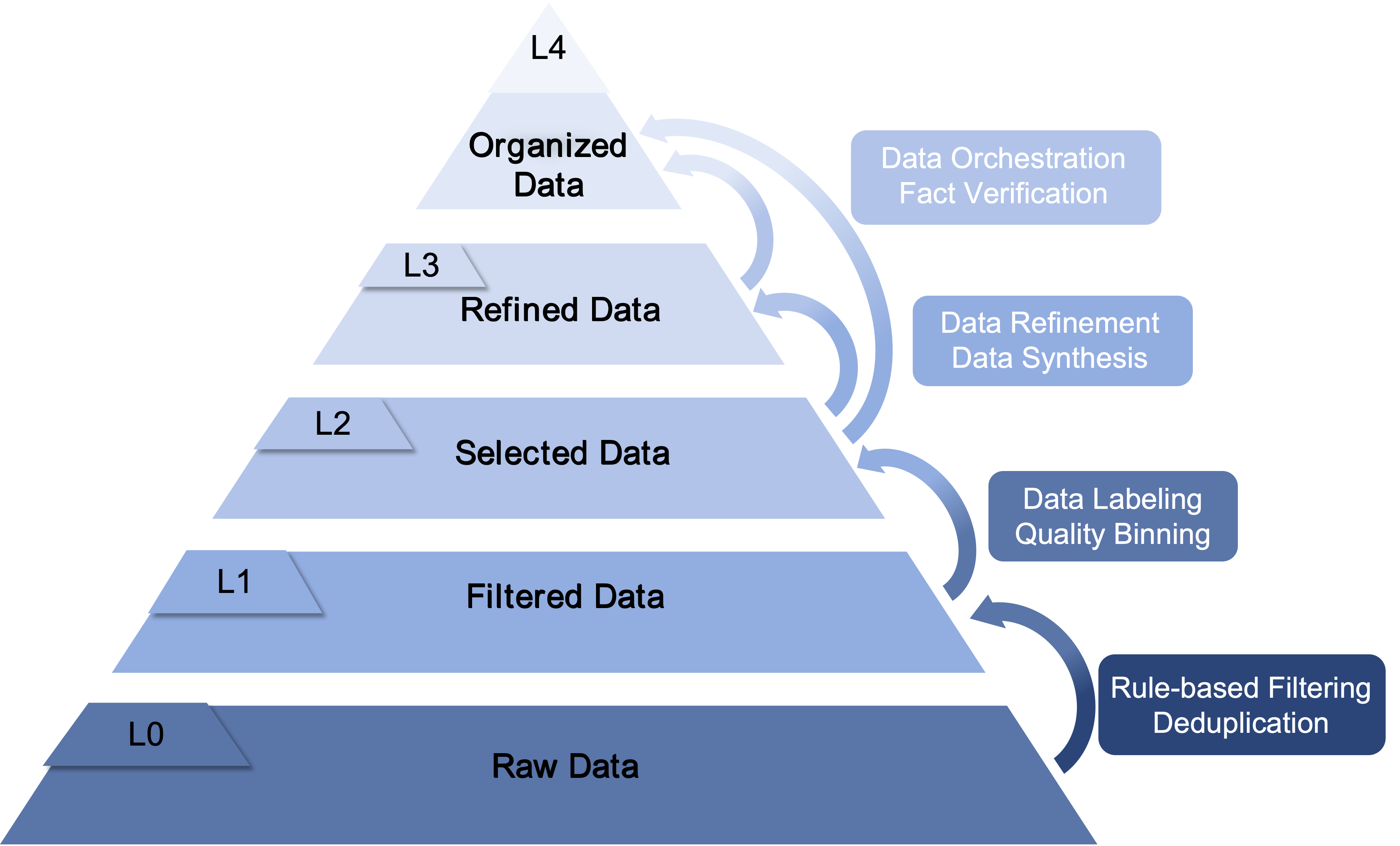}
    \caption{\textbf{Tiered data management framework.} This framework establishes a fine-grained standard centered on data quality and trustworthiness by defining five distinct stages. The framework sequentially elevates data purity through specialized operators while managing the progression from raw data to high-density knowledge assets and their associated computational costs.}
    \label{fig:placeholder}
\end{figure}

% - L0 - 原始数据（Raw Data）
%   - 定义：未经任何实质性处理的初始数据，规模庞大（PB 级），包含大量重复、噪声、广告及无效字符，是治理链路的起点。
%   - 治理手段：数据采集与下载、数据解析。
%   - 代表数据：Common Crawl 网页，PDF 文档、GitHub 仓库
%   - 特征：作为数据仓库的底层储备，信息密度低但覆盖广，用于全量归档与溯源，不直接进入训练管线。
% \textbf{L0: Raw Data}
% \begin{itemize}[leftmargin=1.5em]
%     \item \textbf{Definition:} Raw data without any substantial processing. It is massive in scale (PB-level) and contains massive redundancy, noise, advertisements, and malformed characters.
%     \item \textbf{Operations:} Data Collection, Large-Scale Crawling, and Data Parsing.
%     \item \textbf{Representative Data:} Common Crawl snapshots, raw PDF repositories, and GitHub archives.
%     \item \textbf{Characteristics:} Serves as the foundational reserve of the data warehouse with low information density but broad coverage. It is used for full-scale archiving and provenance tracking rather than direct training.
% \end{itemize}
\subsubsection{L0: Raw Data}
% \vspace{-0.8em}
% L0 层为原始归档数据层，其核心定位是全量数据储备。该层数据以原始采集形态或经基础格式转换后的形式留存，涵盖多类型数据资产：既包括网页通用内容（新闻、博客、论坛讨论、社交媒体、在线百科等）、娱乐内容（影评、游戏攻略、广告推广等）、专业文献（学术论文、技术文档等）、代码仓库资源，也包含音频、视频、图像等多模态数据，同时还夹杂着广告弹窗等各类噪声数据。尽管 L0 层数据的覆盖范围广，但信息密度较低，通常不直接用于模型训练环节。
The L0 level is the raw archival data level, which serves as the foundational reserve of full-scale data. Data at this level is maintained in its native format or undergoes minimal structural conversion. The corpus encompasses a wide array of heterogeneous sources, including general web content (news, blogs, and social media), entertainment media, scholarly literature, and source code repositories. It also integrates multimodal assets such as audio, video, and imagery. Although the L0 level features extensive coverage, its inherent low information density and high noise level render it ineligible for direct use in model training.

% 获取 L0 级原始数据需通过网络爬取、批量下载、格式解析等基础操作实现，针对不同类型的数据源，需适配对应的解析技术工具：面向 HTML 网页解析场景，MinerU-HTML\citep {ma2025aicc} 与 Resiliparse\citep {bevendorff:2018} 可实现高效的 HTML 内容提取；针对 PDF 文档，MinerU\citep {wang2024mineru}、Marker、Nougat\citep {blecher2023nougat} 等工具则能完成从 PDF 到文本数据的精准解析。
Acquisition of L0 data is achieved through basic operations including web crawling, batch downloading and format parsing. For different types of data sources, corresponding parsing technologies and tools are adopted. To facilitate HTML parsing, tools such as MinerU-HTML \citep{ma2025aicc} and Resiliparse \citep{bevendorff:2018} are utilized for efficient content extraction; for PDF documents, tools such as MinerU~\citep{wang2024mineru}, olmOCR~\citep{poznanski2025olmocr} and Nougat~\citep{blecher2023nougat} can accomplish accurate parsing from PDF to textual data.

% 目前具有代表性的 L0 层数据可大约分为三类：在 Web 数据领域，Common Crawl 是规模最大的开放网络爬取数据集，该数据集以快照形式定期发布，最新版本已更新至 2026 年 1 月，包含超 300 亿个网页，时间跨度达 15 年，同时提供原始 WARC 格式（保留完整 HTTP 响应）、HTML 提取后的纯文本 WET 格式等多种版本，可为不同数据治理需求提供灵活选择，而 Mathoverflow 等垂直领域的 Web 数据，也是 L0 层重要的原始数据来源；在学术文献领域，ArXiv 提供预印本论文的 LaTeX 源文件与 PDF 版本；在代码数据领域，GitHub 作为全球最大的代码托管平台，包含数百万个项目的完整代码历史与版本演进记录，Stack Overflow 则提供了丰富的编程问答原始数据。
Currently, representative data at the L0 level are generally classified into three primary categories. In the web data domain, Common Crawl\footnote{\url{https://commoncrawl.org}} is the largest open-source web-crawling dataset, released periodically as snapshot versions. The latest release (updated to January 2026) encompasses over 30 billion web pages spanning 15 years. It provides multiple data granularities for varied data management, including the Web ARChive (WARC) format preserving complete HTTP responses, the Web Archive Transform (WAT) format offering structured metadata and link graphs, and the Web Extracted Text (WET) format providing pre-processed plaintext. In addition, web data from vertical domains such as MathOverflow\footnote{\url{https://mathoverflow.net/}} also acts as a key raw data source for the L0 level. In the academic literature domain, arXiv\footnote{\url{https://arxiv.org/}} provides LaTeX source files and PDF versions of preprint papers. In the code data domain, GitHub\footnote{\url{https://github.com/}}, the world’s largest code hosting platform, houses complete code histories and version evolution records of millions of projects, and Stack Overflow\footnote{\url{https://stackoverflow.com/}} provides abundant raw data for programming Q\&A.

% L0 层的治理重点在于通过数据采集技术实现数据的全量存储，而无需对数据进行额外的质量优化操作，核心目标是保留数据完整性并建立数据溯源基础。由此，可为数据治理体系的后续层级提供充足的原始数据储备，同时赋予整个体系数据回溯能力与二次加工的潜力。
The data management strategy for the L0 level centers on achieving full-scale data storage through data collection technologies, without requiring additional data quality optimization operations. Its core objective is to preserve data integrity and establish a foundation for data traceability. Consequently, this provides abundant raw data reserves for subsequent levels of the data management framework, while endowing the framework itself with data retrospection capabilities and potential for secondary processing.

% - L1 - 过滤数据（Filtered Data）
%   - 定义：经过基础规则清洗的数据。文本格式规范，具备基本的可读性，但语义质量参差不齐。
%   - 治理手段：启发式规则过滤，数据去重、算子编排
%   - 代表数据：FineWeb\citep{penedo2024fineweb}、DCAD\citep{shen2025dcad}
%   - 特征：数据噪声显著减少、基本格式一致性提高、仍保留大部分原始内容；适合作为后续筛选与评估的基础层资源。
% \textbf{L1: Filtered Data} 
% \begin{itemize}[leftmargin=1.5em] 
%     \item \textbf{Definition:} Data with standardized formats and high readability, though semantic quality remains inconsistent and may contain errors (\textit{e.g.}, ad fragments or typos).
%     \item \textbf{Operations:} Rule-based Filtering and Deduplication. 
%     \item \textbf{Representative Data:} FineWeb \citep{penedo2024fineweb} and DCAD \citep{shen2025dcad}. 
%     \item \textbf{Characteristics:} Noise is significantly reduced, and format consistency is improved while retaining most original content. It is suitable as a base resource for subsequent selecting and assessment.
% \end{itemize}
\subsubsection{L1: Filtered Data} 
% \vspace{-0.8em} 
% L1层定义为基础清洗数据层，该层级的核心目标是剔除明显噪声，构建标准化的基础数据集。L1层的治理理念在于通过低成本工程化手段确保数据基本可用性，其技术路径主要包括URL过滤、文本提取、语言识别、启发式规则过滤、全局去重等操作。以下通过RefinedWeb\citep{penedo2023refinedweb}和FineWeb\citep{penedo2024fineweb}两个代表性工作阐述L1层的治理实践。
The L1 level is the basic data-cleaning layer. Its core objective is to eliminate obvious noise and construct a standardized foundational dataset. The data management philosophy for the L1 level centers on ensuring the basic usability of data via low-cost engineering methods, and its technical pipeline primarily comprises operations including URL filtering, text extraction, language identification, heuristic rule filtering, and global deduplication. In what follows, we elaborate on the L1 level data management practices through two representative works: FineWeb~\citep{penedo2024fineweb} and UltraData-Math-L1~\citep{ultradatamath}.

FineWeb \citep{penedo2024fineweb} validates and optimizes L1-level management strategies through empirical ablation experiments. In terms of text extraction, experiments demonstrate that extracting from WARC using \texttt{Trafilatura} significantly improves model performance compared to using WET files. For base filtering, it adopts URL blacklists from RefinedWeb, the \texttt{fastText} language classifier, and quality filtering rules, processing the data down to approximately 36 trillion tokens. Regarding deduplication strategies, findings indicate that global \texttt{MinHash} deduplication increases the sampling rate of low-quality data in older snapshots, leading to performance degradation. Therefore, FineWeb employs an independent snapshot deduplication strategy (5-gram, 112 hash functions, 14 buckets, targeting 75\% similarity), deduplicating each of the 96 Common Crawl snapshots separately to produce about 20 trillion tokens. In terms of heuristic filtering, it integrates rules from the C4 dataset (such as removing lines containing "javascript" or "cookie policy" and filtering overly short documents) and develops custom filters by comparing statistical metrics of high- and low-quality datasets. These custom filters include line-ending punctuation ratio filtering ($\le$ 0.12), duplicated line character ratio filtering ($\ge$ 0.1), and short line ratio filtering ($\ge$ 0.67). Collectively, these filters remove approximately 22\% of tokens and improve the aggregate score by about 1\%. Ultimately, the FineWeb dataset contains 15 trillion tokens and outperforms existing public datasets on multiple benchmarks.

% UltraData-Math\citep{ultradatamath} designs a specialized set of cleaning operators tailored to the unique characteristics of mathematical content for its L1-level management, comprising format repair and content filtering operations. Format repair operators clean and repair data without changing the number of records, specifically including cleaning zero-width characters, invisible spaces, control characters, and garbled characters in Unicode private use areas, cleaning consecutive line breaks, and removing residual noise such as navigation bar text, pagination buttons, and various small buttons; content filtering operators discard entire records that fail to meet requirements, including filtering short articles without punctuation and texts with abnormal lengths. This operator set ensures text standardization through fine-grained format repair while eliminating obviously low-quality samples through content filtering, laying the foundation for subsequent quality filtering and synthesis processing.

UltraData-Math \citep{ultradatamath} implements its L1-level management through a specialized suite of cleaning operators designed for mathematical content. This suite is organized into two primary categories: format repair mappers and content filters. The format repair mappers focus on text standardization without altering the record count. These operators perform fine-grained cleaning tasks, such as eliminating invisible characters and control codes, consolidating excessive consecutive line breaks, and stripping residual interface noise, including navigation bars and pagination buttons. These operations transform raw, noisy extractions into clean, readable text. The content filters focus on record-level pruning based on heuristic rules. Records that fail to meet basic usability standards are discarded entirely. Key filtering criteria include removing short articles that lack proper punctuation and documents with abnormal text lengths. By orchestrating these two types of operators, UltraData-Math-L1 ensures high format consistency and eliminates obviously low-quality samples, establishing a standardized foundation for the subsequent model-driven selection and synthesis stages.

Data at the L1 level is primarily applied to large-scale pre-training, furnishing models with a foundational capacity for general knowledge comprehension and linguistic representation. Marked by low processing costs and high scalability, this level acts as a pivotal stage in constructing high-quality pre-training corpora.

% - L2 - 精筛数据（Selected Data）
%   - 定义：经过价值评估模型筛选，具备高信息密度与特定领域价值的数据，主题明确，逻辑连贯。
%   - 治理手段：模型打分、数据标签
%   - 代表数据：Ultra-FineWeb、FineWeb-edu
%   - 特征：保留对模型能力提升贡献度高的样本，噪声进一步降低，数据密度显著提高，领域明确，信息含量显著提高。
% \textbf{L2: Selected Data} 
% \begin{itemize}[leftmargin=1.5em] 
%     \item \textbf{Definition:} High-value data with clear themes and logical coherence. However, it may still contain minor factual errors, parsing artifacts (\textit{e.g.}, misaligned complex formulas), or subtle gaps in logical derivation.
%     \item \textbf{Operations:} Data Labeling and Quality Binning. 
%     \item \textbf{Representative Data:} Ultra-FineWeb \citep{wang2025ultra} and FineWeb-edu \citep{fineweb}. 
%     \item \textbf{Characteristics:} Retains samples with high contribution to model capability enhancement. It features further reduced noise, significantly increased data density, and clear domain boundaries with high information content.
% \end{itemize}
\subsubsection{L2: Selected Data}
% \vspace{-0.8em} 
The L2 level is defined as the selected data layer, which enhances data information density through model-driven selection mechanisms. This level marks a paradigm shift from rule-based to model-driven data management, with its management strategies comprehensively adopting domain-specific classifiers, semantic-level selecting, quality scoring, data labeling and other approaches. The selection process at this level is essentially a value discovery process; it leverages the discriminative capabilities of models to identify and retain high-value data samples. We elaborate on the L2 level data management practices through two representative works: Ultra-FineWeb \citep{wang2025ultra} and FineMath \citep{allal2025smollm2}.

Ultra-FineWeb \citep{wang2025ultra} proposes a more efficient method for training data selection models, based on the hypothesis that high-quality seed data benefits LLM training and fosters stronger classifiers to identify similar beneficial data. To reduce the prohibitive costs of traditional validation, the method introduces an efficient validation strategy based on a weight-decay scheduler and a two-stage annealing phase. This approach significantly reduces GPU hours required, enabling researchers to rapidly assess the impact of different data subsets. Leveraging this strategy, the researchers rapidly assess the actual impact of different data subsets on LLM performance from a candidate pool, precisely selecting high-quality samples as seed data for a lightweight fastText classifier. Compared to LLM-based classifiers, fastText significantly reduces inference overhead while maintaining high selection quality. Experimental results demonstrate that the Ultra-FineWeb dataset, filtered from FineWeb using this method, significantly outperforms the L1-level FineWeb dataset, validating the effectiveness of model-driven selection over rule-based filtering.

FineMath \citep{allal2025smollm2} addresses the limitations of existing mathematical datasets, such as OpenWebMath \citep{paster2023openwebmath} (14.7B tokens) and InfiMM-WebMath \citep{han2024infimm} (40B tokens), which suffer from insufficient scale and a lack of step-by-step reasoning content. The construction pipeline begins by extracting text from Common Crawl WARC files and using a classifier trained on Llama-3.1-70B-Instruct for initial scoring (on a 3-point scale) to identify high-quality mathematical domains, and then expanding the URL list to include OWM and InfiMM-WebMath. Subsequently, the OWM pipeline is utilized to re-extract all relevant pages, preserving LaTeX formatting while removing boilerplate content, resulting in 7.1B pages and 6.5T tokens. A second classification pass (on a 5-point scale) is then applied to specifically filter for pages containing reasoning and educational content ranging from middle school to early college levels. Finally, the process concludes with single-band MinHash LSH deduplication (10 hashes), fastText language classification (retaining only English), and benchmark decontamination. This yields two versions: FineMath-4+ (10B tokens, retaining scores of 4-5) and FineMath-3+ (34B tokens, retaining scores of 3-5). Experiments show that FineMath-4+ achieves a $2\times$ performance increase on GSM8K and a $6\times$ increase on MATH, significantly outperforming OWM and InfiMM-WebMath.

Compared to L1 data, L2-level data achieves a significant advancement in information density, professionalism, and task relevance. The practices of Ultra-FineWeb and FineMath demonstrate that model-driven quality distillation—whether implemented via classifiers trained on synthetic annotations, fastText models optimized through efficient validation strategies, or multi-stage scoring for specific domains—can effectively extract target-oriented high-quality data assets from general web content. This process provides more efficient training corpora for both foundational pre-training and continuous pre-training stages.

% - L3 - 合成与增强数据（Refined Data）
%   - 定义：经过深度编辑、改写或合成的"完美格式"数据。结构化程度极高，具备逻辑性与教育价值。
%   - 治理手段：数据改写、数据合成、人工标注
%   - 代表数据：Ultra-Chat、UltraFeedback、UltraInteract
%   - 特征：文本可读性强、结构规范、语义清晰，样本质量高，是高阶训练（如 MidTraining、SFT、RL 等）的核心资源。
% \textbf{L3: Refined Data} 
% \begin{itemize}[leftmargin=1.5em] 
%     \item \textbf{Definition:} Highly structured, logical, and “perfectly formatted” data with high educational value. However, it may contain minor hallucinations.
%     \item \textbf{Operations:} Data Refinement and Data Synthesis. 
%     \item \textbf{Representative Data:} UltraChat \citep{ding2023enhancing}, UltraFeedback \citep{cui2023ultrafeedback}, and UltraInteract \citep{yuan2024advancing}. 
%     \item \textbf{Characteristics:} Characterized by strong readability, standardized structure, and clear semantics. This high-quality sample set is the core resource for advanced training stages, such as Mid-training, SFT, and RL.
% \end{itemize}

\begin{table}[htbp]
	\centering
	\caption{Representative open-source tools and datasets for L0--L4 tiered data management.}
	\label{tab:data_tier_system}
	\begin{threeparttable}
	\small
\begin{tabular}{>{\centering\arraybackslash}m{0.6cm} m{13.0cm}}
	\toprule
	\textbf{Level} & \textbf{Open-Source Tools and Datasets} \\
	\midrule
	
\textbf{L0} 
& \textbf{Open-Source Tools}:
\begin{itemize}[nosep, topsep=0pt, partopsep=0pt, itemsep=0pt, parsep=0pt, leftmargin=*]
\item Trafilatura \citep{barbaresi2021trafilatura}, Resiliparse \citep{bevendorff:2018}, MinerU \citep{wang2024mineru}, MinerU-HTML \citep{ma2025aicc}, Nougat \citep{blecher2023nougat}, Docling \citep{auer2024docling}, Mathpix\textsuperscript{\hyperlink{tn:mathpix}{6}}, Magic-HTML\textsuperscript{\hyperlink{tn:magic-html}{7}}, Marker\textsuperscript{\hyperlink{tn:marker}{8}}, olmOCR \citep{poznanski2025olmocr}
\end{itemize}
% \par\vspace{-0.4\baselineskip}\noindent\rule{\linewidth}{0.3pt}\vspace{-0.18\baselineskip}
\textbf{Open-Source Datasets}:
\begin{itemize}[nosep, topsep=0pt, partopsep=0pt, itemsep=0pt, parsep=0pt, leftmargin=*]
\item \textbf{Web}: Common Crawl, Mathoverflow
\item \textbf{PDF}: Papers, Ebooks
\item \textbf{Code}: GitHub, Stackoverflow
\end{itemize}\\
	
	\midrule
	
\textbf{L1} 
& \textbf{Open-Source Tools}:
\begin{itemize}[nosep, topsep=0pt, partopsep=0pt, itemsep=0pt, parsep=0pt, leftmargin=*]
\item MinHash \citep{broder2000identifying}, DataTrove \citep{penedo2024datatrove}, Duplodocus \citep{olmo2025olmo}, Semdedup \citep{abbas2023semdedup}, CCNet \citep{wenzek2020ccnet}
\end{itemize}
% \par\vspace{-0.4\baselineskip}\noindent\rule{\linewidth}{0.3pt}\vspace{-0.18\baselineskip}
\textbf{Open-Source Datasets}:
\begin{itemize}[nosep, topsep=0pt, partopsep=0pt, itemsep=0pt, parsep=0pt, leftmargin=*]
\item \textbf{Web}: C4 \citep{raffel2020exploring}, DCLM-pool \citep{li2024datacomp}, FineWeb \citep{penedo2024fineweb}, FinePDFs \citep{kydlicek2025finepdfs} RefinedWeb \citep{penedo2023refinedweb}, RedPajama-V2 \citep{weber2024redpajama},  RedPajama-V1 \citep{weber2024redpajama}, Dolma \citep{dolma}, WanJuan \citep{he2023wanjuan}, MiChao-HuaFen 1.0 \citep{liu2023michao}, CulturaX \citep{nguyen-etal-2024-culturax}, Txt360 \citep{tang2024txt360}, OSCAR 22.01 \citep{abadji-etal-2022-towards}, SlimPajama \citep{shen2023slimpajama}, CCAligned \citep{el2020ccaligned}, wudao \citep{yuan2021wudaocorpora}, SkyPile-150B \citep{wei2023skywork}
\item \textbf{Code}: The Stack v2 \citep{lozhkov2024starcoder}, The Stack v1 \citep{kocetkov2022stack}, StarCoder \citep{li2023starcoder}, CommitPack \citep{muennighoff2023octopack}, RefineCode \citep{huang2025opencoder}
\item \textbf{Math}: Proof-pile2 \citep{azerbayev2023llemma}, AlgebraicStack \citep{azerbayev2023llemma}
% \item \textbf{Others}: CAIL2018 \citep{xiao2018cail2018}, LEVEN \citep{yao-etal-2022-leven}
\end{itemize}\\
	
	\midrule
	
\textbf{L2} 
& \textbf{Open-Source Tools}:
\begin{itemize}[nosep, topsep=0pt, partopsep=0pt, itemsep=0pt, parsep=0pt, leftmargin=*]
\item Fasttext \citep{joulin2017bag}, Data-Juicer \citep{chen2024data}, Dolma toolkit \citep{dolma}, WebOrganizer \citep{wettig2025organize}
\end{itemize}
% \par\vspace{-0.4\baselineskip}\noindent\rule{\linewidth}{0.3pt}\vspace{-0.18\baselineskip}
\textbf{Open-Source Datasets}:
\begin{itemize}[nosep, topsep=0pt, partopsep=0pt, itemsep=0pt, parsep=0pt, leftmargin=*]
\item \textbf{Web}: DCLM-baseline \citep{li2024datacomp}, FineWeb-Edu \citep{penedo2024fineweb}, FinePDFs-Edu \citep{kydlicek2025finepdfs}, Ultra-FineWeb \citep{wang2025ultra},  IndustryCorpus2 \citep{industryCorpus2},  TeleChat-PTD \citep{wang2024telechat}, CCI3 \citep{wang2024cci30hqlargescalechinesedataset}, ChineseWebText \citep{chinesewebtext}, ChineseWebText2.0 \citep{zhang2024chinesewebtext20largescalehighquality}, Dolma 3 Mix \citep{olmo2025olmo}, Chinese FineWeb-edu \citep{chinese_fineweb}
\item \textbf{Code}: Stack-Edu \citep{allal2025smollm2}, SWE-Gym \citep{pan2024training}, CodeContests \citep{li2022competition}, Apps\citep{hendrycks2021measuring}
\item \textbf{Math}: OpenWebMath \citep{paster2023openwebmath}, FineMath \citep{allal2025smollm2}, Megamath-Web \citep{zhou2025megamath}, MegaMath-Code \citep{zhou2025megamath}, InfiMM-WebMath \cite{han2024infimm}, MathPile \cite{wang2024mathpile}
% \item \textbf{Other}: MegaScience \citep{fan2025megascience} %, SaulLM-7B \citep{colombo2024saullm}
\end{itemize}\\
	
	\midrule
	
\textbf{L3} 
& \textbf{Open-Source Tools}:
\begin{itemize}[nosep, topsep=0pt, partopsep=0pt, itemsep=0pt, parsep=0pt, leftmargin=*]
\item ProX \citep{zhou2024programming}, MoDS \citep{du2023mods}, Self-Instruct \citep{wang2023self}, Evol-Instruct \citep{xu2023wizardlm}, OSS-Instruct \citep{wei2023magicoder}, RLVE \citep{zeng2025rlve}
\end{itemize}
% \par\vspace{-0.4\baselineskip}\noindent\rule{\linewidth}{0.3pt}\vspace{-0.18\baselineskip}
\textbf{Open-Source Datasets}:
\begin{itemize}[nosep, topsep=0pt, partopsep=0pt, itemsep=0pt, parsep=0pt, leftmargin=*]
\item \textbf{Text for pre-training}: Nemotron-CC and Nemotron-CC-HQ \citep{su2025nemotron}, Nemotron-CC-Math-3+/4+ \citep{mahabadi2025nemotron}, MegaMath-Web-Pro \citep{zhou2025megamath}, MegaMath-Synth \citep{zhou2025megamath}, Dolma 3 Dolmino/Longmino Mix and Dolci \citep{olmo2025olmo}
\item \textbf{QA for post-training}: DEITA \citep{liu2023makes}, LIMA \citep{zhou2023lima}, Magpie \citep{xu2024magpie}, UltraFeedBack \citep{cui2024ultrafeedback}, MAmmoTH2 \citep{yue2024mammoth2}, OpenCodeReasoning \citep{ahmad2025opencodereasoning}, OpenThoughts \citep{guha2025openthoughts}, SAND-Math \citep{manem2025sand}, SmolTalk \citep{allal2025smollm2}, Magicoder-OSS-Instruct \citep{wei2023magicoder}, Nemotron-Math-v2 \citep{du2025nemotron}, LIMO \citep{ye2025limo}, LIMO-V2 \citep{ye2025limo}, AugGSM8K and AugMATH \citep{li2024mugglemath}, NuminaMath \citep{li2024numinamath}, Big-Math \citep{albalak2025bigmathlargescalehighqualitymath}, OpenMathInstruct-2 \citep{toshniwal2024openmathinstruct}, Ultrachat \citep{ding2023enhancing}, DISC-Law-SFT \citep{yue2023disclawllm}, SynthLaw \citep{yue2025multi}, FinQA \citep{chen2021finqa}, FinCoT \citep{qian2025fino1}, ConvFinQA \citep{chen2022convfinqa}, SMART’s Trajectory Dataset \citep{yue2024synergistic}, COIG \citep{zhang2023chinese}, OpenCodeInstruct \citep{ahmad2025opencodeinstruct}, WizardCoder \citep{luo2023wizardcoder}, KodCode \citep{xu2025kodcode}, CodeActInstruct \citep{wang2024executable}, SciLitIns \citep{li2024scilitllm}, MegaScience \citep{fan2025megascience}, Skywork-OR1-RL-Data \citep{he2025skywork}
\end{itemize}\\
	
	\midrule
	
\textbf{L4} 
& \textbf{Open-Source Tools}:
\begin{itemize}[nosep, topsep=0pt, partopsep=0pt, itemsep=0pt, parsep=0pt, leftmargin=*]
\item LangChain, LlamaIndex, Haystack, RAGatouille, EmbedChain
\end{itemize}
% \par\vspace{-0.4\baselineskip}\noindent\rule{\linewidth}{0.3pt}\vspace{-0.18\baselineskip}
\textbf{Open-Source Datasets}:
\begin{itemize}[nosep, topsep=0pt, partopsep=0pt, itemsep=0pt, parsep=0pt, leftmargin=*]
\item Wikidata \citep{vrandevcic2014wikidata}, DBpedia, UltraData-arXiv.
\end{itemize}  \\
	
\bottomrule
\end{tabular}
\begin{tablenotes}[para,flushleft]
\footnotesize
\item[6] \hypertarget{tn:mathpix}{} \url{https://mathpix.com/}
\item[7] \hypertarget{tn:magic-html}{} \url{https://github.com/opendatalab/magic-html}
\item[8] \hypertarget{tn:marker}{} \url{https://github.com/datalab-to/marker}
\end{tablenotes}
	\end{threeparttable}
\end{table}

\subsubsection{L3: Refined Data} 
% \vspace{-0.8em} 
% L3 层定义为编辑合成数据层，是在 L1/L2 数据基础上的进阶优化形态。该层级通过编辑修复与合成增强技术消解语义瑕疵，强化逻辑连贯性，为模型能力跃迁提供高阶语料。L3 层标志着数据治理从被动筛选向主动创造的范式跃迁，治理手段涵盖解析后编辑、逻辑修复、合成进化、定制化改写、指令生成等技术。这些方法既优化存量数据质量，又补充增量高质量数据，突破了原始数据在规模和质量上的固有限制。
The L3 level is defined as the edited and synthetic data layer, representing an advanced optimization phase built upon the foundation of L1 and L2 data. This level employs editing, restoration, and synthetic enhancement techniques to eliminate semantic flaws and reinforce logical coherence, providing high-order corpora essential for breakthroughs in model capabilities.

% UltraFineWeb-L3 在通用网络数据上实践了编辑精炼策略。尽管 L2 层的质量分类器已过滤出高分网页，但受解析器准确性限制，这些数据在内容上仍带有不同程度的噪声，样板文本、导航元素、格式不一致等问题普遍存在。借鉴 Nemotron-CC\citep{su2025nemotron}的技术路线，该阶段被视为语义蒸馏过程而非简单过滤，旨在以最纯粹的形式重建底层内容。具体而言，采用 LLM 对筛选后的网页文本进行编辑精炼：系统性移除侧边栏、页眉、页脚和广告残留等非内容元素，修正 OCR 错误、断裂的代码缩进和语法不一致问题，并在不改变原始语义逻辑的前提下提升文本的连贯性和可读性。未能达到最低信息密度阈值的文档将被丢弃。这种"筛选+编辑"范式有效解决了传统基于启发式规则方法的脆弱性问题，实现了数据质量的二次提纯。
UltraFineWeb-L3 \citep{ultradatamath} implements an editing refinement strategy for general web data. Although L2-level quality classifiers have filtered out high-scoring webpages, noise such as boilerplate text, navigation elements, and formatting inconsistencies remains prevalent due to limitations in parser accuracy. Drawing on the technical approach of Nemotron-CC~\citep{su2025nemotron}, this stage treats data processing as semantic distillation rather than simple filtering, aiming to reconstruct the underlying content in its purest form. Specifically, LLMs are employed to refine webpage text by systematically removing non-content elements such as sidebars, headers, footers, and residual advertisements, correcting OCR errors, fixing broken code indentation, and addressing grammatical inconsistencies. This process improves text coherence and readability while strictly preserving the original semantics. Documents failing to meet information density thresholds are discarded, and this "filtering + editing" paradigm effectively overcomes the brittleness of heuristic filtering, achieving secondary purification of data quality.

UltraData-Math-L3~\citep{ultradatamath} implements a systematic synthesis pipeline to overcome the scarcity and homogeneity of web-mined mathematical data. The process begins by cleaning seed documents and standardizing them into a unified LaTeX format, which removes formatting noise and ensures that the model focuses purely on mathematical logic during synthesis. To maximize model generalization, it employs a multi-model ensemble strategy to transform these seeds into five diverse instructional formats: (1) difficulty-stratified Q\&A pairs that follow a curriculum-aligned progression from primary school to undergraduate levels, providing clear supervision signals for problem-solving; (2) multi-turn teacher-student dialogues between seven persona pairs to introduce structural complexity and context-maintenance, which are essential for complex reasoning; (3) multi-style rewrites that decouple mathematical core logic from presentation styles such as Wikipedia, blogs, or academic papers, preventing the model from over-fitting to narrow linguistic distributions; (4) knowledge-driven textbook modules that extract theorems and axioms to generate pedagogical explanations and multi-level practice problems; and (5) persona-integrated synthesis that simulates professional educational materials to further enhance the pedagogical value of the data. Finally, all synthetic outputs undergo rigorous filtering for LaTeX syntax errors and logical incompleteness, ensuring that the final corpus maintains the high information density required for large-scale pre-training.

% L3 层数据适配性强，可支撑模型中期训练、指令微调、强化学习等多个关键阶段，显著提升模型的逻辑推理、数学能力、指令遵循等核心能力。相比 L2 数据，L3 数据不仅密度更高，且通过编辑与合成手段实现了数据价值的创造性增值，突破了原始数据分布在规模、质量和多样性上的固有约束。
L3-level data exhibits strong adaptability, supporting multiple critical stages including mid-training, supervised fine-tuning (SFT), and reinforcement learning (RL), significantly enhancing core model capabilities such as logical reasoning, mathematical proficiency, and instruction following. Compared to L2 data, L3 data not only has higher information density but also achieves creative value augmentation through editing and synthesis, breaking through the inherent constraints of raw data distribution in terms of scale, quality, and diversity.
% - L4 - 编排数据（Organized Data）
%   - 定义：经过统一编排与规范化校验的有序数据。将散乱数据整理成结构清晰、可信任、可检索的数据，可供 RAG 使用。
%   - 治理手段：数据编排、可信校验
%   - 代表数据：wikidata\citep{vrandevcic2014wikidata}、Ultra-arXiv
%   - 特征：数据具备清晰结构、可检索性与高可信度。
% \textbf{L4: Organized Data}
% \begin{itemize}[leftmargin=1.5em] 
%     \item \textbf{Definition:} Organized data that has undergone unified orchestration and rigorous trustworthiness verification. It transforms scattered information into structured, reliable, and searchable assets.
%     \item \textbf{Operations:} Data Orchestration and Fact Verification. 
%     \item \textbf{Representative Data:} Wikidata \citep{vrandevcic2014wikidata} and UltraData-arXiv. 
%     \item \textbf{Characteristics:} This level represents data with explicit structure and high credibility. It facilitates efficient retrieval, such as for Retrieval-Augmented Generation (RAG), and ensures the factual integrity required for knowledge-intensive applications. 
% \end{itemize}
\subsubsection{L4: Organized Data} 
% \vspace{-0.8em} 
The L4 level is defined as the organized data layer, representing the most refined state of data within the management framework. While previous levels focus on the linguistic and semantic quality of continuous text, the L4 level emphasizes unified orchestration and rigorous normalization to transform fragmented information into structured, reliable, and searchable knowledge assets. This layer serves as the authoritative ``source of truth'' for models, providing the high-fidelity substrate necessary for knowledge-intensive tasks.

The management of L4 data centers on two core operations: data orchestration and fact verification. Through orchestration, scattered data from diverse sources are unified under coherent thematic frameworks and interconnected knowledge structures. Simultaneously, fact verification ensures the integrity of the information by cross-referencing entries with trusted sources, thereby eliminating the factual inconsistencies often found in raw web corpora.

Representative examples of L4 data include Wikidata \citep{vrandevcic2014wikidata} and UltraData-arXiv. Wikidata provides a highly structured, multilingual knowledge base that facilitates precise entity-relation querying. Similarly, Ultra-arXiv represents a sophisticated reorganization of scholarly literature, where complex elements such as mathematical formulas, citations, and experimental results are standardized into a searchable and interconnected format.

The defining characteristics of the L4 level are its explicit structural rigor and exceptional credibility. These attributes make L4 data indispensable for advanced applications like retrieval-augmented generation (RAG). By enabling efficient and accurate retrieval, L4 data provides a robust defense against model hallucinations and ensures the factual precision required for expert-level reasoning and decision-making.

% 通过构建 L0–L4 的精细化分级治理体系，数据治理实现了从“粗放式堆量”向“精细化赋能”的范式转型。该体系支撑了渐进式训练策略，使模型能够在不同演进阶段精准匹配质量递增的数据资源。L0 通常作为全量归档的储备数据，不参与到模型的实际训练。通过通用清洗规则和定制化算子处理后的 L1 数据可支撑大规模预训练。L2 通过分类模型可以筛选出信息密度高，满足不同领域和质量需求的数据，适用于 Decay 与 MidTraining 阶段。在此基础上，通过改写与合成进一步消解语义瑕疵的 L3 数据，为模型逻辑能力跃迁提供高阶语料，适用于 MidTraining、SFT 和 RL 多个阶段。而 L4（编排数据）则通过规范化校验和数据编排构建可信的知识索引，可直接服务于 RAG 等下游应用。该分级体系有效地解耦了数据治理的复杂挑战，避免了传统治理中“一刀切”的低效，实现了更严谨的质量控制。这种结构为下游适配提供了模块化资源库，允许研究者依据特定需求从 $L1$ 池中进行领域定向抽样，或提取 $L2$ 优质种子样本以指导 $L3$ 专用语料的进化合成。配合 $L4$ 层级构建的可信知识索引，该体系将数据质量从模糊的经验评价转化为可预测的工程指标，从而在可追溯的数据血缘基础上，驱动模型能力实现从经验驱动向价值驱动的确定性演进。
By establishing the fine-grained, hierarchical L0–L4 management framework, data management undergoes a paradigm shift from ``extensive accumulation'' to ``precise empowerment''. This framework underpins a progressive training strategy, enabling models to be precisely matched with data resources of incrementally improving quality at different stages of evolution.
L0 data generally serves as archival reserve material and does not participate in actual model training. L1 data, processed through general cleaning rules and customized operators, can support large-scale pre-training. L2 data, filtered via classification models to possess high information density and meet diverse domain and quality requirements, is suitable for the Decay and MidTraining stages. Building upon this, L3 data, further refined through rewriting and synthesis to resolve semantic imperfections, provides high quality corpus for leaps in model logical reasoning capabilities, applicable across MidTraining, SFT, and RL stages. Meanwhile, L4 (Organized Data), through standardized verification and data orchestration, constructs a trustworthy knowledge index that can directly serve downstream applications such as RAG.
This hierarchical framework effectively decouples the complex challenges of data management, avoiding the inefficiency of a traditional "one-size-fits-all" approach and enabling more rigorous quality control. Its structured framework provides a modular resource library for downstream adaptation, allowing researchers to perform domain-targeted sampling from the L1 pool based on specific needs, or extract high-quality seed samples from L2 to guide the evolutionary synthesis of dedicated L3 corpora. Complemented by the trusted knowledge index constructed at the L4 level, this framework transforms data quality from vague, experience-based assessments into predictable engineering metrics. Consequently, building upon traceable data lineage, it drives model capabilities toward deterministic advancement, shifting from an experience-driven to a value-driven paradigm.

\section{Experiments}
In this section, we present a comprehensive evaluation of our proposed tiered data management framework.
We first detail the experimental setting in Section~\ref{exp:exp_set}, including model configuration, various verification strategies (pre-training, efficient, and decay verification), and evaluation benchmarks.
In Section~\ref{exp:data_analysis}, we perform a granular quality analysis across four major corpora: English Web, Chinese Web, Math, and Code. Our results demonstrate that data quality improves progressively from L1 to L3, validating the effectiveness of the tiered management in enhancing corpus quality and purity.
Subsequently, Section~\ref{exp:math_exp} provides a specialized case study on mathematics to systematically validate the L1–L3 framework. 
% We demonstrate that leveraging UltraData-Math-L3 leads to state-of-the-art (SOTA) performance, providing a definitive benchmark for the effectiveness of our tiered data management framework.
% 
Finally, Section~\ref{exp:tierred_train} presents a case study on tiered management for model training. By exploring the deployment of different data tiers across various training stages, we reveal that higher-level data becomes increasingly critical as training progresses.

\subsection{Experimental Setting}
\label{exp:exp_set}
\textbf{Model Configuration.} In our experiments, all models are trained via the Megatron-LM library~\citep{megatron}. We utilize the MiniCPM-1.2B model architecture with the MiniCPM3-4B tokenizer. Table \ref{tab:model_cfg} provides the detailed configurations, where {Params.}, {Vocab.}, $d_m$, $d_{ff}$, $d_h$, $n_{head}$, $n_{kv}$, and $n_{Layer}$ represent the total number of non-embedding parameters, vocabulary size, model hidden dimension, feedforward layer bottleneck dimension, attention head dimension, number of queries, number of key/values, and the number of layers, respectively.

% Table Model CFG
\begin{table}[!h]
    \centering
    \small
    \caption{Model configurations for the MiniCPM-1.2B.}
    \renewcommand{\arraystretch}{1.2}
    \begin{tabular}{ccccccccc}
        \toprule
        Name & Params. & Vocab. & $d_m$ & $d_{ff}$ & $d_h$ & $n_{head}$ & $n_{kv}$ & $n_{Layer}$ \\
        \midrule
        MiniCPM-1.2B & 1,247,442,432 & 73448 & 1,536 & 3,840 & 64 & 24 & 8 & 52 \\
        \bottomrule
    \end{tabular}
    \label{tab:model_cfg}
\end{table}

\textbf{Efficient Verification.}
To efficiently evaluate data quality, we conduct a two-stage annealing process on a 10B token budget. 
We first train a base model from scratch on 1.1T tokens using the MiniCPM-3-4B corpus. 
The training employs a Warmup-Stable-Decay (WSD) scheduler, comprising a 1T-token stable stage and a 0.1T-token decay stage. 
Building upon this base, we perform annealing with a data mixture composed of 30\% verification data and 70\% of the default distribution. 
Key training parameters include a sequence length of 4096, weight decay of 0.1, and a gradient clipping threshold of 1.0. We employ a global batch size of 512 (micro batch size of 16).
This configuration yields a total of 10.5B tokens, calculated as $\textit{SeqLen}~\times~\textit{GBS}~\times~\textit{TrainStep}~= 4096 \times 512 \times 5000 = 10.5$B; for simplicity, we refer to it as 10B. 
The optimization for this stage employs an exponential decay schedule with a 500-step warm-up, scaling the learning rate from a peak of $1 \times 10^{-3}$ to a minimum of $5 \times 10^{-5}$. To enhance training stability, we use Maximal Update Parameterization ($\mu$P)~\citep{mup} across all experimental setups.

\textbf{Pre-train Verification.}
We perform pre-training verification on approximately 120B tokens to balance validation comprehensiveness with computational efficiency. 
Each verification comprises 15,000 steps with a sequence length of 4,096 and a global batch size of 2,048 (implemented with a micro-batch size of 16). 
This configuration yields a total of 125.8B tokens, calculated as $\textit{SeqLen}~\times~\textit{GBS}~\times~\textit{TrainStep}~=~4096\times 2048 \times 15000 = 125.8$B tokens; for simplicity, we refer to it as $120$B.
The optimization process employs a cosine learning rate decay schedule with a 1,000-step warm-up phase. The learning rate initiates at $1 \times 10^{-5}$, reaches a peak of $1 \times 10^{-2}$, and subsequently decays to $5 \times 10^{-4}$. 
Additional hyperparameters include a weight decay of 0.1 and a gradient clipping threshold of 1.0. Similarly, $\mu$P is employed to improve the stability of the training process.

\textbf{Decay Verification.}
To evaluate data performance during the final pre-train phase, we conduct decay verification using a base model trained on 1.3T tokens from the MiniCPM-4 corpus, having completed both warmup and stable stages.
For this verification, we employ a data mixture composed of 30\% new verification data and 70\% of the default distribution. 
The process involves 20,000 steps with a sequence length of 4,096 and a global batch size of 1,280 (micro-batch size of 10), yielding approximately 104.9B tokens ($4096 \times 1280 \times 20000$), which we refer to as 100B for brevity.
We utilize an exponential decay schedule that scales the learning rate from the stable stage's $7.5 \times 10^{-4}$ down to a minimum of $3.75 \times 10^{-5}$. Other training hyperparameters remain consistent with our standard settings. Although efficient verification enables higher iteration efficiency with significantly lower resource requirements, the limited training budget may introduce higher variance in results. Decay verification addresses this by employing a full-scale decay phase, providing a more robust and definitive evaluation that closely reflects the final pre-training performance.

\textbf{Benchmarks.} We adopt OpenCompass~\citep{opencompass} as our evaluation framework. 
The specific benchmarks, evaluation settings, and inference methods are detailed as follows:
\begin{itemize}[leftmargin=*]
    % - 英文平均：MMLU、ARC-C、ARC-E、CommonSenseQA、HellaSwag、OpenbookQA、PIQA、SIQA 和 Winograd
    \item General English datasets: We evaluate our models on standard English benchmarks including MMLU (5-shot, PPL)~\citep{mmlu}, ARC-C (0-shot, PPL)~\citep{arc}, ARC-E (0-shot, PPL)~\citep{arc}, BigBench Hard~(BBH) (3-shot, Gen)~\citep{bbh}, CommonSenseQA (8-shot, PPL)~\citep{commonsenseqa}, HellaSwag (0-shot, PPL)~\citep{hellaswag}, OpenbookQA (0-shot, PPL)~\citep{openbookqa} (0-shot, PPL), PIQA (0-shot, PPL)~\citep{piqa} (0-shot, PPL), SIQA (0-shot, PPL)~\citep{siqa} (0-shot, PPL), and Winogrande (0-shot, Loglikelihood)~\citep{winogrande}.
    % - 中文平均：C-Eval 和 CMMLU
    \item General Chinese datasets: We evaluate our models on Chinese knowledge-intensive benchmarks, including C-Eval (5-shot, PPL)~\citep{ceval} and CMMLU (5-shot, PPL)~\citep{cmmlu}.
    \item Math reasoning datasets: We evaluate our models on mathematics reasoning benchmarks, including MATH500 (4-shot, Gen)~\citep{hendrycks2measuring} and GSM8K~\citep{cobbe2021training} (4-shot, Gen).
    \item Code reasoning datasets: We evaluate our models on code reasoning benchmarks, including MBPP (3-shot, Gen)~\citep{austin2021program} and HumanEval (0-shot, Gen)~\citep{chen2021evaluating}.
    % - 总平均：上述所有评估指标的综合平均得分。
    % \item \textit{Average}: The combined average score of all the above evaluation benchmarks.
\end{itemize}

\subsection{Data Analysis}
\label{exp:data_analysis}
% 快速验证
% 分别对比 web-en、web-zh、code、math、mix
% web-en: fineweb、ultra-fineweb-en、ultra-fineweb-en-l3
% web-zh: Chinese fineweb、ultra-fineweb-zh、ultra-fineweb-zh-l3
% code: stack-v2、stack-edu、stack-textbook-gen
% math: ultradata-math-l1、ultradata-math-l2、ultradata-math-l3、
% 快速验证
\begin{table}[!t]
    \centering
    \caption{Detailed data selection for $L1, L2,$ and $L3$ across multiple domains.}
    \label{tab:data_selection}
    \resizebox{\columnwidth}{!}{ % 强制缩放到栏宽
    \renewcommand{\arraystretch}{1.2}
    \begin{tabular}{l|lll}
        \toprule
        Domain & L1 Filtered Data & L2 Selected Data & L3 Refined Data \\
        \midrule
        Web-en & FineWeb~\citep{fineweb} & Ultra-FineWeb-en~\citep{wang2025ultra} & Ultra-FineWeb-en-L3 \\
        Web-zh & Chinese FineWeb~\citep{chinese_fineweb} & Ultra-FineWeb-zh~\citep{wang2025ultra} & Ultra-FineWeb-zh-L3 \\
        Math   & UD-Math-L1~\citep{ultradatamath} & UD-Math-L2~\citep{ultradatamath} & UD-Math-L3~\citep{ultradatamath} \\
        Code   & Stack-v2~\citep{lozhkov2024starcoder} & Stack-Edu~\citep{allal2025smollm2} & Code Textbook \\
        \bottomrule
    \end{tabular}
    }
\end{table}

We first conduct an efficient verification method to confirm that the tiered data management framework produces meaningful quality differentiation. Specifically, we evaluate the data quality across the L1, L2, and L3 tiers for English web, Chinese web, Math, and Code domains. For each domain, the tiered composition is defined as follows. 
(1)~In the English web domain, we select FineWeb~\citep{fineweb} as L1 data, Ultra-FineWeb-en~\citep{wang2025ultra} as L2, and Ultra-FineWeb-en-L3, a synthetic dataset generated based on Ultra-FineWeb-en as L3. Following prior synthetic data construction practices (e.g., Nemotron-CC~\citep{su2025nemotron}), the L3 data comprises five types of synthesized content spanning diverse supervision signals (\textit{i.e.}, diverse QA, distill, extract knowledge, knowledge list, and wiki style). 
(2)~Similarly, for the Chinese web domain, Chinese FineWeb (source data from Chinese FineWeb-edu-v2~\citep{chinese_fineweb}) is defined as L1, Ultra-FineWeb-zh~\citep{wang2025ultra} as L2, and Ultra-FineWeb-zh-L3, which is constructed via multi-type synthetic data generation on top of Ultra-FineWeb-zh as L3.
(3)~For the math domain, we uniformly adopt the UltraData-Math~\citep{ultradatamath} for data construction and management across all tiers. Specifically, Math-L1 consists of raw mathematical text parsed from web data using the UltraData-Math Parser and filtered with rule-based filter operators. Math-L2 is obtained by further selecting high-quality samples using the UltraData-Math Classifier, while Math-L3 is composed of synthetic mathematical data generated by the UltraData-Math Generator.
(4)~In the code domain, we select Stack-v2~\citep{lozhkov2024starcoder} as Code-L1 and Stack-Edu~\citep{allal2025smollm2} as Code-L2, while Code-L3 is derived from Code-L2 through textbook-style rewriting, including code explanations and programming exercises.
We apply the same efficient verification strategy across all domains and tiers to ensure fair and comparable quality assessment. The experimental results are shown in Table~\ref{tab:level_results}, demonstrating a clear and consistent performance improvement from L1 to L3 across domains, validating that the proposed tiered data management framework effectively captures meaningful quality stratification.

\begin{table*}[!t]
    \centering
    \small
    \caption{Comparison of models trained with L1, L2, and L3 datasets across English web, Chinese web, math, and code domains.}
    \renewcommand{\arraystretch}{1.2}

    \begin{tabular*}{\textwidth}{l @{\extracolsep{\fill}} cccccccccc c}
        \toprule
        \multirow{2}{*}[-2pt]{\makecell[l]{Web-En\\Data}} & \multicolumn{11}{c}{English} \\
        \cmidrule(lr){2-12}
        & MMLU & ARC-E & ARC-C & BBH & CSQA & Hella. & OBQA & PIQA & SIQA & Wino. & Avg. \\
        \midrule
        L1 & 46.88 & \textbf{61.38} & 37.63 & \textbf{35.36} & 57.82 & 56.87 & 56.00 & 72.85 & 42.94 & 54.85 & 52.26 \\
        L2 & 46.73 & 59.79 & \textbf{39.32} & 34.94 & 57.99 & 56.72 & 66.20 & \textbf{73.29} & 43.30 & 55.33 & 53.36 \\
        L3 & \textbf{47.25} & 59.08 & 37.97 & 34.51 & \textbf{58.56} & \textbf{57.12} & \textbf{72.40} & 73.18 & \textbf{43.40} & \textbf{56.12} & \textbf{53.96} \\
        \bottomrule
    \end{tabular*}

    % \vspace{-2pt}

    \begin{tabular*}{\textwidth}{l @{\extracolsep{\fill}} ccc | l @{\extracolsep{\fill}} ccc | l @{\extracolsep{\fill}} ccc}
        \toprule
        \multirow{2}{*}[-2pt]{\makecell[l]{Web-Zh\\Data}} & \multicolumn{3}{c|}{Chinese} & \multirow{2}{*}[-2pt]{\makecell[l]{Math\\Data}} & \multicolumn{3}{c|}{Math} & \multirow{2}{*}[-2pt]{\makecell[l]{Code\\Data}} & \multicolumn{3}{c}{Code} \\
        \cmidrule(lr){2-4} \cmidrule(lr){6-8} \cmidrule(lr){10-12}
        & CMMLU & C-Eval & Avg. & & MATH500 & GSM8K & Avg. & & MBPP & HumanEval & Avg. \\
        \midrule
        L1 & 49.37 & 49.51 & 49.44 & L1 & 14.80 & 32.75 & 23.78 & L1 & 43.97 & 25.00 & 34.49 \\
        L2 & 50.72 & 50.59 & 50.66 & L2 & 15.60 & 34.04 & 24.82 & L2 & 44.36 & 26.22 & 35.29 \\
        L3 & \textbf{51.74} & \textbf{51.22} & \textbf{51.48} & L3 & \textbf{20.20} & \textbf{41.47} & \textbf{30.84} & L3 & \textbf{45.73} & \textbf{26.83} & \textbf{36.28} \\
        \bottomrule
    \end{tabular*}
    \label{tab:level_results}
    \vspace{-1.5em}
\end{table*}

Across all four domains, downstream performance improves steadily from L1 to L3, demonstrating that data quality increases with each data tier. Specifically, average benchmark scores rise from 52.26 percentage points (pp) to 53.96pp in English (+1.70 pp), 49.44pp to 51.48pp in Chinese (+2.04 pp), 23.78pp to 30.84pp in Math (+7.06 pp), and 34.49pp to 36.28pp in Code (+1.79 pp). The strict performance hierarchy of $L3 > L2 > L1$ holds universally without exception. These results empirically validate the effectiveness of our tiered data management framework in producing high-quality data, which yields clear performance signals even under verification with constrained training budgets.

\subsection{Case Study on UltraData-Math}
\label{exp:math_exp}
% To further examine whether quality improvements in a single domain can translate into broad downstream gains, we scale up to 100B tokens using our UltraData-Math corpus. We train models using math-domain data at each quality level (L1, L2, L3) and evaluate across all benchmarks, spanning not only mathematical reasoning but also English understanding and code generation. Table\ref{tab:math_level_result} reports the full results.
To further examine whether quality improvements in a single domain can translate into broad downstream gains, we scale up to 100B tokens using the UltraData-Math corpus. 
This scaling experiment aims to verify whether the quality advantages of tiered data are scalable, ensuring sustained performance gains without premature saturation as the training volume increases.
Specifically, we utilize the decay verification method to verify whether the quality-driven dividends observed during small-scale verification persist and scale linearly as the training trajectory extends.
By training models separately on UltraData-Math-L1, UltraData-Math-L2, and UltraData-Math-L3 data, we evaluate their performance across a diverse suite of benchmarks spanning mathematical reasoning, English and Chinese understanding, and code generation. 

\begin{table*}[!b]
    \centering
    \small
    \caption{Comparison of results across Math data quality levels (L1--L3) on all benchmarks.}
    \renewcommand{\arraystretch}{1.2}

    \begin{tabular*}{\textwidth}{l @{\extracolsep{\fill}} cccccccccc c}
        \toprule
        \multirow{2}{*}{Method} & \multicolumn{11}{c}{English} \\
        \cmidrule(lr){2-12}
        & MMLU & ARC-E & ARC-C & BBH & CSQA & Hella. & OBQA & PIQA & SIQA & Wino. & Avg. \\
        \midrule
        Math-L1 & 50.57 & 54.50 & 37.29 & 37.75 & 60.44 & 58.02 & 41.60 & 74.21 & 41.71 & 57.14 & 51.32 \\
        Math-L2 & 50.93 & 55.20 & 36.95 & 39.27 & 60.20 & 57.52 & 39.80 & 74.48 & \textbf{44.73} & 57.77 & 51.69 \\
        Math-L3 & \textbf{51.67} & \textbf{59.79} & \textbf{38.98} & \textbf{43.62} & \textbf{61.18} & \textbf{58.27} & \textbf{57.00} & \textbf{74.76} & 43.35 & \textbf{59.04} & \textbf{54.77} \\
        \bottomrule
    \end{tabular*}

    % \vspace{-2pt}

    \begin{tabular*}{\textwidth}{l @{\extracolsep{\fill}} ccc ccc ccc c}
        \toprule
        \multirow{2}{*}{Method} & \multicolumn{3}{c}{Chinese} & \multicolumn{3}{c}{Math} & \multicolumn{3}{c}{Code} & \multirow{2}{*}{All Avg.} \\
        \cmidrule(lr){2-4} \cmidrule(lr){5-7} \cmidrule(lr){8-10}
        & CMMLU & C-Eval & Avg. & MATH500 & GSM8K & \textit{Avg.} & MBPP & HumanEval & Avg. & \\
        \midrule
        Math-L1 & 51.28 & 51.89 & 51.59 & 27.78 & 54.66 & 41.22 & 44.71 & 29.88 & 37.30 & 48.39 \\
        Math-L2 & 51.13 & 50.55 & 50.84 & 29.20 & 52.92 & 41.06 & 44.50 & 32.32 & 38.41 & 48.59 \\
        Math-L3 & \textbf{52.87} & \textbf{54.08} & \textbf{53.48} & \textbf{37.02} & \textbf{61.79} & \textbf{49.41} & \textbf{49.27} & \textbf{32.93} & \textbf{41.10} & \textbf{58.27} \\
        \bottomrule
    \end{tabular*}
    \label{tab:math_level_result}
\end{table*}

Furthermore, the benefits of Math-L3 extend beyond mathematical reasoning and translate into consistent improvements across non-mathematical domains. On English benchmarks, Math-L3 improves the average score from 51.32pp and 51.69pp to 54.77pp, yielding gains of 3.45pp and 3.08pp over Math-L1 and Math-L2, respectively. Notably, substantial improvements are observed on reasoning-intensive tasks such as ARC-E (+5.29pp), ARC-C (+1.69pp), BBH (+5.87pp), and OpenbookQA (+15.40pp), indicating enhanced general reasoning and problem-solving capabilities induced by higher-quality math data. Similar trends are observed in the Chinese domain, where Math-L3 achieves an average score of 53.48pp, outperforming Math-L1 and Math-L2 by 1.89pp and 2.64pp, respectively, with consistent gains on both CMMLU and C-Eval. In the code domain, Math-L3 also demonstrates clear advantages, improving the average performance to 41.10pp, compared to 37.30pp and 38.41pp for Math-L1 and Math-L2. This improvement is reflected across both MBPP (+4.56pp over Math-L1) and HumanEval (+3.05pp over Math-L1), suggesting that high-quality mathematical data can enhance structured reasoning and abstraction skills beneficial to code generation tasks. Overall, these results demonstrate that progressively improving data quality within a single domain can yield significant and transferable benefits across diverse evaluation domains. These cross-domain improvements underscore that high-quality mathematical data is a fundamental driver of enhancing a model's general logical consistency and problem-solving capabilities across diverse languages and tasks.

\subsection{Tiered Data Management for Multi-Stage Training}
\label{exp:tierred_train}
To validate the effectiveness of the tiered data management framework, we design a comparative experiment to evaluate the performance of mix training and tiered training strategies under the same training setting. 
Both training strategies adhere to a consistent domain distribution consisting of 50\% Web-en, 25\% Web-zh, 8\% Math, and 17\% Code.
The mix training strategy utilizes a single-stage approach, mixing 120B tokens with an equal 1:1:1 ratio of L1, L2, and L3 data into a unified pool.
In contrast, the tiered training strategy partitions the same 120B tokens into three consecutive 40B token stages (effectively equivalent to a 1:1:1 ratio), transitioning from L1 to L2 and finally to L3 data.
Both strategies employ the MiniCPM-1.2B model trained from scratch and are evaluated on multiple benchmarks across four major domains: English, Chinese, Math, and Code.

\begin{table*}[!t]
    \centering
    \small
    \caption{Comparison of mix training and tiered training results across all benchmarks.}
    \renewcommand{\arraystretch}{1.2}

    \begin{tabular*}{\textwidth}{l @{\extracolsep{\fill}} cccccccccc c}
        \toprule
        \multirow{2}{*}{Method} & \multicolumn{11}{c}{\textit{English}} \\
        \cmidrule(lr){2-12}
        & MMLU & ARC-E & ARC-C & BBH & CSQA & Hella. & OBQA & PIQA & SIQA & Wino. & Avg. \\
        \midrule
        Mix    & 28.26 & 48.32 & 26.78 & 26.20 & 46.11 & \textbf{46.89} & 26.00 & \textbf{71.44} & \textbf{39.76} & \textbf{54.30} & 41.41 \\
        Tiered & \textbf{29.15} & \textbf{50.09} & \textbf{31.53} & \textbf{28.37} & \textbf{46.27} & 45.21 & \textbf{29.00} & 70.62 & 39.20 & 53.43 & \textbf{42.29} \\
        \bottomrule
    \end{tabular*}

    % \vspace{-2pt}

    \begin{tabular*}{\textwidth}{l @{\extracolsep{\fill}} ccc ccc ccc c}
        \toprule
        \multirow{2}{*}{Method} & \multicolumn{3}{c}{Chinese} & \multicolumn{3}{c}{Math} & \multicolumn{3}{c}{Code} & \multirow{2}{*}{All Avg.} \\
        \cmidrule(lr){2-4} \cmidrule(lr){5-7} \cmidrule(lr){8-10}
        & CMMLU & C-Eval & \textit{Avg.} & MATH500 & GSM8K & \textit{Avg.} & MBPP & HumanEval & \textit{Avg.} & \\
        \midrule
        Mix    & 25.47 & 23.97 & 24.72 & 1.60 & 3.11 & 2.36 & 12.06 & 2.44 & 7.25 & 30.17 \\
        Tiered & \textbf{26.71} & \textbf{28.37} & \textbf{27.54} & \textbf{4.20} & \textbf{5.00} & \textbf{4.60} & \textbf{16.34} & \textbf{3.05} & \textbf{9.70} & \textbf{31.66} \\
        \bottomrule
    \end{tabular*}
    \label{tab:main_result}
    \vspace{-1em}
\end{table*}

\begin{wrapfigure}{r}{0.55\textwidth}
    \centering
    \includegraphics[width=0.5\textwidth]{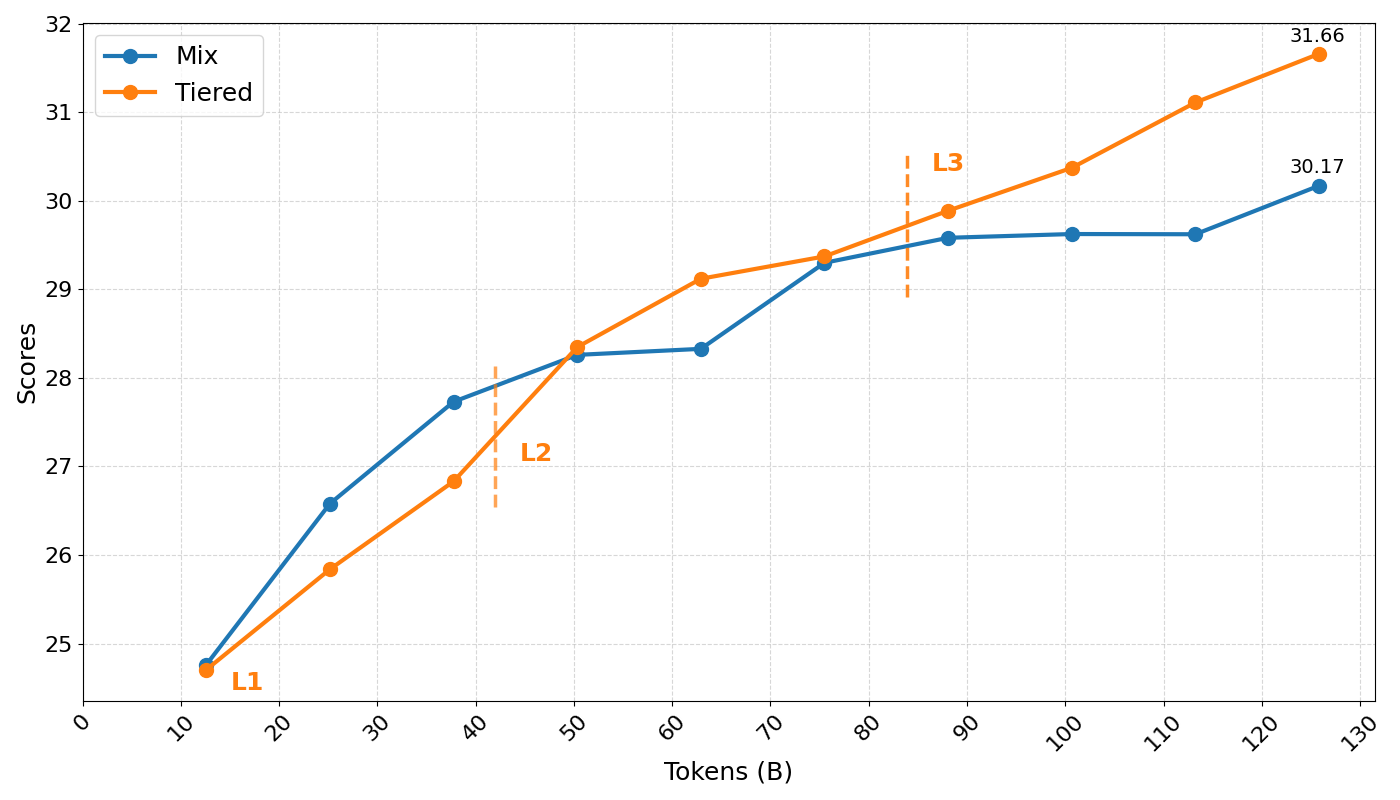}
    % \caption{Comparison of mix training and tiered training average scores at each checkpoint.}
    \caption{\textbf{Comparison of average score between mix training and tiered training at each checkpoint.} The tiered training strategy (40B tokens per stage) consistently outperforms the mix training baseline across most evaluation intervals.}
    \label{fig:mix_tiered_fig}
\end{wrapfigure}
Table~\ref{tab:main_result} presents detailed comparative results of the two strategies across evaluation benchmarks. Tiered training achieves an improvement of 1.49pp in overall average performance compared to mix training (31.66pp vs 30.17pp), with significant gains across all four major evaluation domains. Specifically, the English domain shows an average improvement of 0.88pp, with reasoning-intensive tasks such as ARC-C, ARC-E, OpenbookQA, and BBH improving by 4.75pp, 1.77pp, 3.00pp, and 2.17pp respectively, and knowledge-intensive tasks such as MMLU improving by 0.89pp; the Chinese domain shows an average improvement of 2.82pp, with C-Eval and CMMLU improving by 4.40pp and 1.24pp respectively; the mathematics domain shows an average improvement of 2.24pp, with MATH500 and GSM8K improving by 2.60pp and 1.89pp respectively; the code domain shows an average improvement of 2.45pp, with MBPP and HumanEval improving by 4.28pp and 0.61pp respectively. 
In-depth analysis reveals that tiered training demonstrates more significant improvements on reasoning-intensive tasks (ARC-C, BBH, OpenbookQA) and knowledge-intensive tasks (MMLU, C-Eval, CMMLU), benefiting from the model-driven selection mechanism of the L2 that effectively identifies and retains high information density samples, while the editing and synthesis techniques of the L3 further enhance the logical coherence of data. The substantial improvements in the math and code domains validate the effectiveness of domain-specific management strategies: through L2's domain classifier selection and L3's editing enhancement, the model can more fully absorb domain knowledge. It is worth noting that tiered training shows slight decreases on a few tasks (HellaSwag, PIQA, SIQA, Winogrande), which focus more on the breadth rather than depth of common sense reasoning and language understanding, potentially benefiting from the larger scale L1 data coverage in mix training.

Figure~\ref{fig:mix_tiered_fig} further reveals the advantages of tiered management from the dynamic perspective of the training process. In the early training stage, both strategies exhibit similar growth trends, with performance improving from approximately 24.7pp to around 28.3pp. 
At this stage, tiered training primarily uses L1 data, with a data distribution relatively close to mix training.
In the latter stages of training, the tiered training strategy gradually introduces high-quality data from L2 and L3, and the performance curve demonstrates a sustained and stable growth trend, improving from 28.35pp to 31.66pp, an increase of 3.31pp.
In contrast, the growth trend of mix training significantly slows down, improving from 28.26pp to 30.17pp, an increase of only approximately 1.91pp.
This significant difference fully embodies the core advantage of tiered governance: by introducing high-quality data that has undergone L2-model-driven selection and L3-editing and synthesis optimization in the later training stages, the model can continuously and efficiently learn complex knowledge and reasoning capabilities, avoiding the learning efficiency decline caused by low-quality data interference in mix training.

% \begin{figure}[ht]
%     \centering
%     \includegraphics[width=0.8\textwidth]{figs/mix_tiered_img.png}
%     % \caption{Comparison of mix training and tiered training average scores at each checkpoint.}
%     \caption{\textbf{Comparison of average score between mix training and tiered training at each checkpoint.} The tiered training strategy (40B tokens per stage) consistently outperforms the mix training baseline across most evaluation intervals.}
%     \label{fig:mix_tiered_fig}
% \end{figure}

Combining results on benchmarks and training curves, this experiment fully validates the effectiveness of the tiered data management framework. 
The tiered training strategy not only comprehensively surpasses mix training in final performance (overall average improvement of 1.49pp), but more importantly, demonstrates a sustained and stable trend of learning capability improvement in the later training stages, with a growth magnitude reaching 1.7 times that of the mix training strategy. 
This advantage stems from the core mechanism of tiered management: by introducing data of corresponding quality levels at different training stages, achieving precise alignment between data value and model learning demands. 
L1 data provides a broad foundation of language representation in the early training stage, L2 data enhances the learning of high information density content in the mid-stage, and L3 data deepens the absorption of logical reasoning and domain knowledge in the final stage, thereby effectively avoiding the interference of low-quality data in the mix training strategy on the learning of advanced capabilities. 
The experimental results demonstrate that organizing 120B tokens of training data into tiered stages according to L1, L2, and L3 can more effectively enhance the model's comprehensive performance across multiple dimensions, such as knowledge understanding, logical reasoning, and domain capabilities, providing solid empirical support for the stepwise optimization of data management.

\section{Conclusion}
In this paper, we revisit the development of artificial intelligence through the lens of data organization and utilization, and argue that the dominant paradigm of data-driven learning is approaching fundamental sustainability limits. As model capabilities continue to advance, further progress can no longer rely solely on expanding data scale, but instead requires a systematic rethinking of how data is managed, valued, and deployed throughout the training lifecycle.
To this end, we propose a data-model co-evolution perspective, in which models actively guide data management decisions while high-quality data, in turn, amplifies model capability in a positive feedback loop. Under this paradigm, we introduce a L0–L4 tiered data management framework that structures data from raw resources to organized and verifiable knowledge. By explicitly aligning data quality, management cost, and training objectives across different learning stages, the proposed framework provides a principled and scalable foundation for sustainable LLM data management.
Through empirical studies on math and web data, we demonstrate that tier-aware data utilization can significantly improve training efficiency and model performance, validating the practical value of tiered data management beyond isolated data processing techniques. Our results suggest that effective data management should be treated as a first-class engineering problem, rather than an auxiliary preprocessing step.

Our future work will focus on deepening and operationalizing the data-model co-evolution paradigm. Specifically, we plan to develop more rigorous methods for scientific data value assessment, enabling models to quantitatively estimate the marginal utility of data across tiers and training stages. We will further explore dynamic data–model feedback mechanisms, where model signals continuously inform data selection, refinement, and allocation during training. In addition, we aim to extend the tiered data management framework to broader modalities and application domains, and to integrate it more tightly with large-scale training systems. Through these efforts, we seek to establish data management as a core, adaptive component of next-generation AI systems.

\newpage

\bibliographystyle{citation}
\bibliography{citation}

\end{document}